\documentclass{article}

\usepackage[final]{neurips_2025}

\usepackage[utf8]{inputenc} % allow utf-8 input
\usepackage[T1]{fontenc}    % use 8-bit T1 fonts
\usepackage[colorlinks=true, linkcolor=BrickRed, citecolor=NavyBlue, urlcolor=NavyBlue]{hyperref}
\usepackage{url}            % simple URL typesetting
\usepackage{booktabs}       % professional-quality tables
\usepackage{amsfonts}       % blackboard math symbols
\usepackage{nicefrac}       % compact symbols for 1/2, etc.
\usepackage{microtype}      % microtypography
\usepackage[table, dvipsnames]{xcolor}     % colors
\usepackage{enumitem}
\usepackage{xspace}
\usepackage{amsmath}
\usepackage{graphicx}
\usepackage{mathtools}

\usepackage{multirow}
\usepackage{soul}
\usepackage{changepage,threeparttable}
\usepackage{subcaption}
\usepackage{wrapfig}
\usepackage[export]{adjustbox}
\newcommand{\summary}[1]{}

\newcommand{\method}{\textit{UniAug}\xspace}

\author{
Wenzhuo Tang\textsuperscript{1},
Haitao Mao\textsuperscript{2}\thanks{This work was conducted while H.M. was a Ph.D. student at Michigan State University and is independent of H.M.’s current position at Amazon.} ,
Danial Dervovic\textsuperscript{3},
Ivan Brugere\textsuperscript{3},
Saumitra Mishra\textsuperscript{3},\\
\textbf{Yuying Xie\textsuperscript{1}},
\textbf{Jiliang Tang\textsuperscript{1}}\\
\textsuperscript{1}\,Michigan State University \quad
\textsuperscript{2}\,Amazon \quad
\textsuperscript{3}\,J.P. Morgan AI Research \quad
\\
\texttt{\{tangwen2, xyy, tangjili\}@msu.edu,\ maoht@amazon.com}\\
\texttt{\{danial.dervovic, ivan.brugere, saumitra.mishra\}@jpmchase.com}
}

\title{Cross-Domain Graph Data Scaling: \\A Showcase with Diffusion Models}

\begin{document}

\maketitle

\begin{abstract}
    % Large-scale models have shown great potential in many domains. 
% The effectiveness of foundation models largely relies on the availability of substantial amounts of data.
% As the scale of pre-training data increases, the capabilities of these models grow accordingly, exemplifying data scaling behavior.
% However, different graph tasks have different inductive biases, making transferability on graphs challenging. 
% Consequently, existing graph pre-training methods do not capitalize on data scaling, resulting in the underutilization of many unlabeled datasets.
% Conceptually, the pre-training process should not lead to worse performance and may find better patterns with more data.
% Following the spirit, 
% In this work, we propose to separate the upstream and downstream processes by utilizing graph structure augmentation. 
% We train a foundation graph structure augmentor, \method, on thousands of graphs from various domains.
% In the downstream phase, we perform adaptive augmentation for task-specific models to enable positive transfer.
% Empirically, we apply \method to graphs from diverse domains and consistently observe performance improvements in node classification, link prediction, and graph classification. 
% We further examine the scaling behavior of \method and investigate potential negative transfers from the augmentation process.
% To the best of our knowledge, this study represents the first demonstration of a foundation graph structure augmentor on graphs across domains.

% \summary{what is data scaling} 
Models for natural language and images benefit from data scaling behavior: the more data fed into the model, the better they perform.
% \summary{data scaling enables training on large-scale data} 
This 'better with more' phenomenon enables the effectiveness of large-scale pre-training on vast amounts of data.
% \summary{large-scale data -> model with general knowledge and adaptive on downstream}
% With the help of large-scale data, 
% fine-tune it to benefit different downstream tasks.
% \summary{graph does not benefit from data scaling}
However, current graph pre-training methods struggle to scale up data due to heterogeneity 
% \wenzhuo{wording for 'heterogeneity'} 
across graphs. 
% \summary{model name}
To achieve effective data scaling, we aim to develop a general model that is able to capture diverse data patterns of graphs and can be utilized to adaptively help the downstream tasks.
% Consequently, we are able to develop a model with general knowledge that adaptively supports downstream tasks for natural language and images.
To this end, we propose \method, a universal graph structure augmentor built on a diffusion model.
% \summary{diffusion model for general knowledge}
We first pre-train a discrete diffusion model on thousands of graphs across domains to learn the graph structural patterns.
% \summary{adaptive on downstream: use it as an augmentor}
In the downstream phase, we provide adaptive enhancement by conducting graph structure augmentation with the help of the pre-trained diffusion model via guided generation.
% \summary{universal, plu-and-playable}
By leveraging the pre-trained diffusion model for structure augmentation, we consistently achieve performance improvements across various downstream tasks in a plug-and-play manner.
% model architectures for downstream tasks 
% \summary{statement}
To the best of our knowledge, this study represents the first demonstration of a data-scaling graph structure augmentor on graphs across domains.
% Our implementations are publicly available on \href{https://github.com/WenzhuoTang/UniAug}{https://github.com/WenzhuoTang/UniAug}.

% \vspace{-1.5em}
\end{abstract}

\section{Introduction}

The effectiveness of existing foundation models~\citep{radford2021learning, touvron2023llama, kirillov2023segment} heavily relies on the availability of substantial amounts of data, where the relationship manifests as a scaling behavior between model performance and data scale~\citep{kaplan2020scaling}.
\summary{NLP, CV models benefit from data scaling} 
Consistent performance gain has been observed with the increasing scale of pre-training data in both Natural Language Processing~\citep{kaplan2020scaling, hoffmann2022training} and Computer Vision~\citep{abnar2022exploring, zhai2022scaling} domains.
% With the increase in the magnitude of pre-training data, we observe the growth of model capability for natural language~\citep{kaplan2020scaling, hoffmann2022training} and images~\citep{abnar2022exploring, zhai2022scaling}. 
% \mao{the success of LLM, LVM, multimodal. model scaling and data scaling all important, data scaling }
\summary{large-scale data -> model with general knowledge} 
This data scaling phenomenon facilitates the development of general models endowed with extensive knowledge and effective data pattern recognition capabilities.
% enormous knowledge and effective data pattern.
% This 'better with more' phenomenon facilitates the development of general models endowed with universal knowledge, \mao{I cannot understand the following ones}which serve as a powerful prior when combined with large-scale pre-training on vast amounts of data.
\summary{large-scale data -> adaptive on downstream} 
In downstream applications, these models are capable of adaptively achieving performance gains across tasks.
% with minimal data-specific modifications to model parameters.

% \summary{graph has many databases -> possible data scaling} 
In the context of graphs, the availability of large-scale graph databases~\citep{rossi2015network, hu2020OGB, snapnets} enables possible data scaling across datasets and domains.
% \mao{enables to increase data scaling}
% \wenzhuo{put this paragraph before the above one}
% \summary{limited solution: domain-/task-specific data selection} 
Existing works have demonstrated graph data scaling following two limited settings: in-domain pre-training~\citep{xia2023molebert, liu2024rethinking} and 
% \jt{it is unclear "task-specific selection for pre-training data", maybe just "task-specific pretraining"??}
task-specific selection for pre-training data~\citep{cao2023pre}.
% two paths towards graph data scaling: 
% A possible solution to eliminate the negative transfer is manually selecting pre-training data, including in-domain pre-training and adaptive~\citep{cao2023pre}. 
% \summary{drawback: manually select relevant graphs}
% \jt{maybe we should not say manually validate, the data we use also via some selection process. I think their data is limited to specific domains or specific tasks instead of cross tasks and cross domains. That is sufficient to show their limitations}
During the pre-training process, each graph in the pre-training pool must be 
% \jt{I still think manually here is not good, maybe we can remove "manually", maybe validated itself} 
% manually 
validated as in-domain or relevant to the downstream dataset.
% This is because different downstream tasks in different domains require specific inductive biases.
% resulting in loss of potentially useful data.
% many graphs being discarded. 
% \summary{drawback: 2. limited scaling} This filtering procedure can lead to the loss of potentially useful data, thereby only offering limited data scaling benefits.
% \summary{drawback: 1. not general: limited generalizability} These methods are typically designed for a single domain or task, which limits their generalizability.
% \summary{drawback: 1. not general: one model per task/domain} Specifically, they require a unique data collection and pre-training phase for each domain or task, coupled with a model structure tailored accordingly, leading to significant resource consumption when applied to multiple tasks or domains.
% \summary{drawback: 2. manually select part of the data}
% In addition, during the pre-training process, each graph in the pre-training pool must be manually validated as in-domain or relevant to the downstream dataset, resulting in many graphs being discarded. 
% \summary{because: 3. distinct inductive biases require specific designs} Another contributing factor for the failure of data scaling lies in the diverse inductive biases characterized by different downstream tasks, which necessitate task-specific model designs~\citep{mao2024demystifying, mao2024revisiting, xu2018powerful}.
Given a specific domain or task, the crucial discriminative data patterns are likely confined to a fixed set~\citep{mao2024revisiting}, leaving other potential patterns in diverse graph data distribution as noisy input.
In terms of structure, graphs from different domains are particularly composed of varied patterns~\citep{milo2002network}, making it hard to transfer across domains.
% could potentially share a common set of network motifs, which are small and recurring subgraphs serving as the fundamental building blocks of a graph~\citep{milo2002network}
For example, 
% \jt{any references for the following claim??} 
considering the building blocks of the graphs, the motifs shared by the World Wide Web hyperlinks only partially align with those shared by genetic networks~\citep{milo2002network}.
Therefore, closely aligning the characteristics of the pre-training graphs and the downstream data both in feature and structure is essential for facilitating positive transfer~\citep{cao2023pre}.
% \mao{Given the specific domain or task, the crucial discriminative data patterns are likely to be fixed~\citep{mao2024revisiting}, leaving other potential patterns in diverse graph data distribution as noisy patterns.
% Closely matching the characteristics~\citep{cao2023pre} of the pre-training graphs and the downstream data both in feature and structure is necceary for positive transfer.}
% \summary{why only part of the data? pre-training data should be close to downstream data} 
% This filtering process is crucial for graph pre-training methods, as it enables transferability by requiring that the pre-training graphs should closely match the characteristics of the downstream data both in feature and structure~\citep{cao2023pre}.
% \summary{drawback: limited data scaling}
As a consequence, the necessity of such meticulous data filtering restricts these methods from scaling up graphs effectively, as they can only utilize a small part of the available data.
% \summary{current pre-training does not benefit from data scaling} However, current methods for graph pre-training suffer from the 'curse of big data' \wenzhuo{call back to data scaling} phenomenon, where large pre-training datasets may diminish model performance~\citep{xu2023better, cao2023pre}.
% \summary{question} 
Given the limitation of the graph pre-training methods, a pertinent question emerges: \textit{How can we effectively leverage the increasing scale of graph data across domains?}

Rather than focusing solely on data patterns specific to particular domains, we aim to develop a model that has a comprehensive understanding of data patterns inherent across various types of graphs. 
In line with the principles of data scaling, we hypothesize that incorporating a broader range of training datasets can help the model build an effective and universal graph pattern library,
% \jt{if we use "vocabulary" here, we need to explicitly define what it means in our work???} 
% pattern base \wenzhuo{wording for 'base'}, 
avoiding an overemphasis on major data patterns specific to any single dataset~\citep{mao2024demystifying}.
To construct such a general-purpose model, we propose to utilize a diffusion model operating only on the structure as the backbone, for the following key reasons.
(1) Unlike features, graph structures follow a uniform construction principle, namely, the connections between nodes. 
This allows for positive transfer across domains when the upstream and downstream data exhibit similar topological patterns~\citep{cao2023pre}. 
In particular, while the graph representations of neurons and forward electronic circuits are derived from distinct domains, they still share common motifs~\citep{milo2002network}.
(2) Current supervised and self-supervised methods tend to capture only specific patterns of graph data, with models designed for particular inductive biases~\citep{mao2024demystifying, mao2024revisiting, xu2018powerful}.
For instance, graph convolutional networks (GCNs) excel in node-level representation learning by emphasizing homophily, whereas graph-level representation learning benefits from expressive GNNs capable of distinguishing complex graph structures.
(3) We opt for a structure-only model due to the heterogeneous feature spaces across graphs, which often include missing features or mismatched semantics~\citep{mao2024graph}. 
For instance, node features yield completely different interpretations in citation networks, where they represent keywords of documents, compared to molecular networks, where they denote properties of atoms.
To this end, we pre-train a structure-only diffusion model on thousands of graphs, which serves as the upstream component of our framework.

% Across data scaling requiring no specific graph inductive bias design
% Another contributing factor for the failure of data scaling lies in the diverse inductive biases characterized by different downstream tasks, which necessitate task-specific model designs~\citep{mao2024demystifying, mao2024revisiting, xu2018powerful}.
% For instance, while graph convolutional networks (GCNs) excel in node classification, they perform poorly in graph-level tasks compared to more expressive GNNs. 
% This makes it extremely challenging to build a pre-training and fine-tuning pipeline with a single model backbone to obtain performance improvement across tasks.

% \summary{our expectation: a general model for universal understanding}
% In line with the principles of data scaling, we hypothesize that incorporating a broader range of training datasets may enable models to develop a comprehensive understanding of the data patterns inherent in graphs.
% \summary{our expectation: a general model that can help different tasks} 
% Specifically, the pre-trained model should be free of any downstream-specific inductive biases and should enhance performance across various downstream tasks.
% \summary{our solution: train a diffusion model on various graphs} 
% To achieve this goal, we propose to pre-train a self-conditioned discrete diffusion model on thousands of unlabeled graphs to capture diverse structure patterns of graphs.
% \summary{our solution: do adaptive augmentation for downstream} 
In the downstream stage, we employ the pre-trained diffusion model as a \textbf{Uni}versal graph structure \textbf{Aug}mentor (\textbf{UniAug}) to enhance the dataset, where diffusion guidance~\citep{ho2022classifier, dhariwal2021diffusion, gruver2024protein} is employed to align the generated structure with the downstream requirements.
% \jt{in the previous sections, we list challenges from three perspectives? when we discuss the advantages of our methods, can we disucss how our method solves the previous three challenges???}
% \summary{augmented graph: generated structures and original node features} 
Specifically, we generate synthetic structures with various guidance objectives, and the resulting graphs consist of \textit{generated structures} and \textit{original node features}.
% \summary{strength of our solution: plug-and-play}
This data augmentation paradigm strategically circumvents feature heterogeneity and fully utilizes downstream inductive biases by applying carefully designed downstream models to the augmented graphs in a plug-and-play manner.
% This data augmentation paradigm allows \method to seamlessly leverage the capabilities of both the pre-trained diffusion model and the carefully designed downstream models in a plug-and-playable manner.
% \summary{results in performance boost} 
Empirically, we apply \method to graphs from diverse domains and consistently observe performance improvement in node classification, link prediction, and graph property prediction. 
% \summary{statement} 
To the best of our knowledge, this study represents the first demonstration of a cross-domain data-scaling graph structure augmentor.

\section{Preliminary and Related Work}
% In this section, we introduce the related works and backgrounds of discrete diffusion models on graph adjacency. We summarize the comparison between methods in Table~\ref{tab: method_comp}.
% \vspace{-0.5em}
% \input{tables/method_comp}

\label{sec: background}

\paragraph{Learning from unlabeled graphs.} 
Graph self-supervised learning (SSL) methods provide examples of pre-training and fine-tuning paradigm~\citep{hu2019strategies, hou2022graphmae, kim2022graph, you2021graph, xu2021self}. 
However, these methods benefit from limited data scaling due to feature heterogeneity, structural pattern differences across domains, and varying downstream inductive biases.
It is worth mentioning that DCT~\citep{liu2024data} presents a pre-training and then data augmentation pipeline on molecules. Despite its impressive performance improvement on graph-level tasks, DCT is bounded with molecules and thus the use cases are limited.

% \textbf{Graph foundation models}

\paragraph{Graph data augmentation.} 
There have been many published works exploring graph data augmentation (GDA) since the introduction of graph neural networks (GNNs), with a focus on node-level~\citep{park2021metropolis, liu2022local, azabou2023half}, link-level~\citep{zhao2022learning, nguyen2024diffusion}, and graph-level~\citep{han2022g, ling2023graph, luo2022automated, liu2022graph, kong2022robust}. 
These GDA methods have been generally designed for specific tasks or particular aspects of graph data. 
In addition, they are often tailored for a single dataset and struggle to transfer to unseen patterns, which limits their generalizability to a broader class of applications.

\paragraph{Diffusion models on graphs.} 
% \wenzhuo{high-level introduction of how diffusion models work}
Diffusion models~\citep{ho2020denoising, song2021scorebased, rombach2022high} are latent variable models that learn data distribution by gradually adding noise into the data and then recovering the clean input.
Existing diffusion models on graphs can be classified into two main categories depending on the type of noise injected, i.e. Gaussian or discrete. Previous works employed Gaussian diffusion models both on general graphs~\citep{niu2020permutation, jo2022score} and molecules~\citep{shi2021learning, xu2022geodiff}. However, adding Gaussian noise into the adjacency matrix will destroy the sparsity of the graph, which hinders the scalability of the diffusion models~\citep{haefeli2022diffusion}. Recent works adapted discrete diffusion models to graphs with categorical transition kernels~\citep{vignac2023digress, chen2023efficient, chen2023dexplainer}. 
% In the following, we will formulate the existing discrete diffusion models into binary diffusion on the adjacency matrix.
We denote the adjacency matrix of a graph as $\mathbf{A}^0 \in \{0, 1\}^{n\times n}$ with $n$ nodes.
% and denote one element of $\mathbf{A}^0$ as $\mathbf{A}^0$. 
% Following D3PM~\citep{austin2021structured}, we corrupt the adjacency matrix into a sequence of latent variables $\mathbf{A}^{1:T} = \mathbf{A}^1, \mathbf{A}^2, \dots, \mathbf{A}^T$ by independently injecting noise into each element with a Markov process
% \begin{equation}
    % q\left(\mathbf{A}^t \mid \mathbf{A}^{t-1}\right)=\prod_{i,j: i<j} \operatorname{Cat}\left(\mathbf{A}^t_{ij} ; \mathbf{p}=\mathbf{A}^{t-1}_{ij} \mathbf{Q}^t \right),
% \end{equation}
% where $\mathbf{Q}^t \in [0, 1]^{2\times 2}$ is the transition probability of timestep $t$. The above Markov process is called \textit{forward process}. With some derivations, 
With details in Appendix~\ref{app: diffusion}, we write the \textit{forward process} to corrupt the adjacency matrix into a sequence of latent variables as Bernoulli distribution
\vspace{0.25em}
\begin{equation}
% \vspace{-0.2pt}
% \begin{split}
% \resizebox{0.4\textwidth}{!}{$
\begin{aligned}
    &q\left(\mathbf{A}^t \mid \mathbf{A}^{t-1}\right) =\operatorname{Bernoulli}\left(\mathbf{A}^t ; \alpha^t \mathbf{A}^{t-1} + \left(1-\alpha^{t}\right) \pi \right), \\
    % q\left(\mathbf{A}^t \mid \mathbf{A}^{0}\right) & =\operatorname{Bernoulli}\left(\mathbf{A}^t  ; \bar{\alpha}^{t} \mathbf{A}^0 +\left(1-\bar{\alpha}^{t}\right) \pi \right), \\
    &q\left(\mathbf{A}^{t-1} \mid \mathbf{A}^{t}, \mathbf{A}^{0}\right) = \frac{q\left(\mathbf{A}^{t} \mid \mathbf{A}^{t-1}\right) q\left(\mathbf{A}^{t-1} \mid \mathbf{A}^{0}\right)}{q\left(\mathbf{A}^{t} \mid \mathbf{A}^{0}\right)},
\end{aligned}
% $}
% \end{split}
\end{equation}
\vspace{0.25em}
where $\pi$ is the converging non-zero probability, $\alpha^t$ is the noise scale, and $\bar{\alpha}^{t} = \prod_{i=1}^t \alpha^i$. 
% The main difference of the \textit{forward process} among the existing works is the choice of prior $\pi$, where $\pi=0$ for EDGE~\citep{chen2023efficient}, $\pi=0.5$ for D4Explainer~\citep{chen2023dexplainer}, and a pre-computed $\pi$ for DiGress~\citep{vignac2023digress}. 
Under predict-$\mathbf{A}^0$ parameterization, the \textit{reverse process} denoise the adjacency matrix with a Markov chain
% \vspace{-0.5em}
\vspace{0.25em}
\begin{equation}
% \vspace{-0.5em}
% \resizebox{0.6\textwidth}{!}{$
    p_{\theta}\left(\mathbf{A}^{t-1} \mid \mathbf{A}^{t}\right) \propto \sum_{\widetilde{\mathbf{A}}_{0}} q\left(\mathbf{A}^{t-1} \mid \mathbf{A}^{t}, \widetilde{\mathbf{A}}_{0}\right) \tilde{p}_{\theta}\left(\widetilde{\mathbf{A}}_{0} \mid \mathbf{A}^{t}\right),
% $}
\end{equation}
\vspace{0.25em}
% \vspace{-0.5em}
where $\tilde{p}_{\theta}(\widetilde{\mathbf{A}}_{0} \mid \mathbf{A}^{t})$ represents the denoising network that predicts the original adjacency matrix from the noisy adjacency matrix. 
The parameters are estimated by optimizing the variational lower bound on the negative log-likelihood~\citep{austin2021structured}
% \vspace{-0.5em}
\vspace{0.25em}
\begin{equation}
\label{eq: vlb}
% \vspace{-5pt}
% \begin{split}
% \resizebox{\textwidth}{!}{$
    % L_{\mathrm{vb}}= \sum_{t=2}^{T} \mathbb{E}_{q\left(\mathbf{A}^{t} \mid \mathbf{A}^{0}\right)}\left[D_{\mathrm{KL}}\left(q\left(\mathbf{A}^{t-1} \mid \mathbf{A}^{t}, \mathbf{A}^{0}\right)\Vert\ p_{\theta}\left(\mathbf{A}^{t-1} \mid \mathbf{A}^{t}\right)\right)\right]  - \mathbb{E}_{q\left(\mathbf{A}^{1} \mid \mathbf{A}^{0}\right)}\left[\log p_{\theta}\left(\mathbf{A}^{0} \mid \mathbf{A}^{1}\right)\right] + \mathbb{E}_{q\left(\mathbf{A}^{0}\right)} \left[D_{\mathrm{KL}}\left(q\left(\mathbf{A}^{t} \mid \mathbf{A}^{0}\right) \Vert\ p\left(\mathbf{A}^{t}\right)\right)\right] .
\begin{aligned}
    L_{\mathrm{vb}}= & \sum_{t=2}^{T} \mathbb{E}_{q\left(\mathbf{A}^{t} \mid \mathbf{A}^{0}\right)}\left[D_{\mathrm{KL}}\left(q\left(\mathbf{A}^{t-1} \mid \mathbf{A}^{t}, \mathbf{A}^{0}\right)\Vert\ p_{\theta}\left(\mathbf{A}^{t-1} \mid \mathbf{A}^{t}\right)\right)\right] \\
    & - \mathbb{E}_{q\left(\mathbf{A}^{1} \mid \mathbf{A}^{0}\right)}\left[\log p_{\theta}\left(\mathbf{A}^{0} \mid \mathbf{A}^{1}\right)\right] + \mathbb{E}_{q\left(\mathbf{A}^{0}\right)} \left[D_{\mathrm{KL}}\left(q\left(\mathbf{A}^{t} \mid \mathbf{A}^{0}\right) \Vert\ p\left(\mathbf{A}^{t}\right)\right)\right] .
\end{aligned}
% $}
% \end{split}
\end{equation}

% \vspace{-1.5em}

\section{Method}
\label{sec: method}

% \vspace{-0.5em}

% \jt{for the followign subsections, we present what we have done but we need present why first and then how.  We need to motivate why we choose data augmentation. I think let us split the framework into pretraining stage and then downstream stage??  Even though we may discuss these challenges in the intro, feel free to recall or even elabrate them in the methods part }

\begin{figure*}[!tp]
    \centering
    % \vspace{-2em}
    % \resizebox{\textwidth}{!}{
    \includegraphics[width=0.95\textwidth]{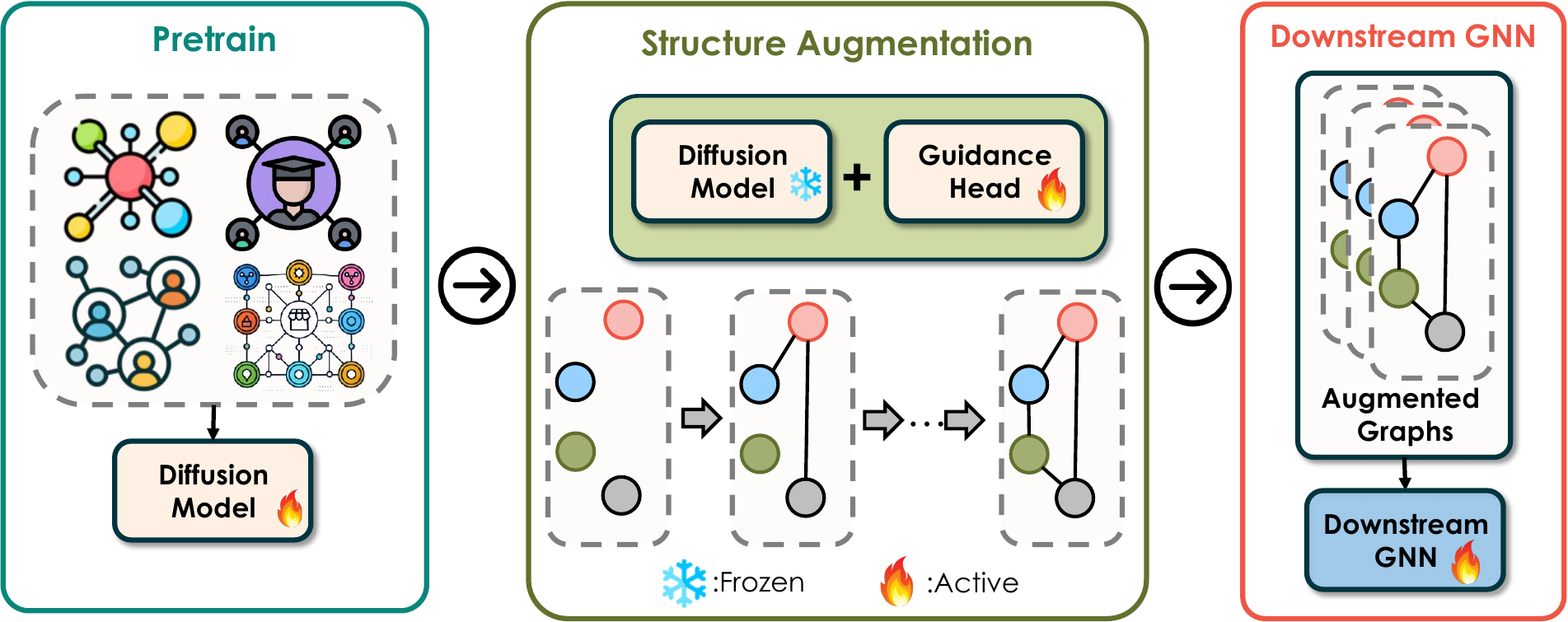}
    % }
    % \vspace{0.5em}
    \caption{
    % \small
    The pipeline of \method. 
    We pre-train a diffusion model across domains and perform structure augmentation on the downstream graphs.
    The augmented graphs consist of \textit{generated structures} and \textit{original node features} and are then processed by a downstream GNN.
    }
    % \vspace{-0.5em}
    \label{fig: framework}
\end{figure*}

% \vspace{-0.5em}

In this section, our goal is to build \method to understand the diverse structure patterns of graphs and perform data augmentation with a range of objectives. 
% \jt{I think we still need to recall the challenges we need to solve to achieve this goal?  Then we present oru framework and high-levely describe how our design solves these challenges?? I think our framework design is not well motivated}
As illustrated in Fig.\ref{fig: framework}, \method consists of two main components: a pre-trained diffusion model and the downstream adaptation through structure augmentation.
% \wenzhuo{add motivation: diffusion model is more general without inductive bias}
We first collect thousands of graphs from varied domains with diverse patterns.
To construct a general model free of downstream inductive biases, we train a self-conditioned discrete diffusion model on graph structures. 
In the downstream stage, we train an \textit{MLP guidance head} on top of the diffusion model with objectives across different levels of granularity.
We then augment the downstream dataset by generating synthetic structures through guided generation, where the augmented graph is composed of \textit{generated structures} and \textit{original node features}.
Subsequently, we apply the augmented data to train a task-specific model for performing downstream tasks. 
Below, we elaborate on the data collection process, the architecture of the discrete diffusion model, and the guidance objectives. 
% Details of the data collection process can be found in Appendix~\ref{app: datasets}.

% \vspace{-1em}

\subsection{Pre-training data collection}
% \vspace{-0.5em}

\label{sec: pre_data}

\begin{wrapfigure}{r}{0.50\textwidth} % {0.4\textwidth}
% \begin{figure}[!htp]
    \vspace{-1em}
    \begin{center}
        \includegraphics[width=0.40\textwidth]{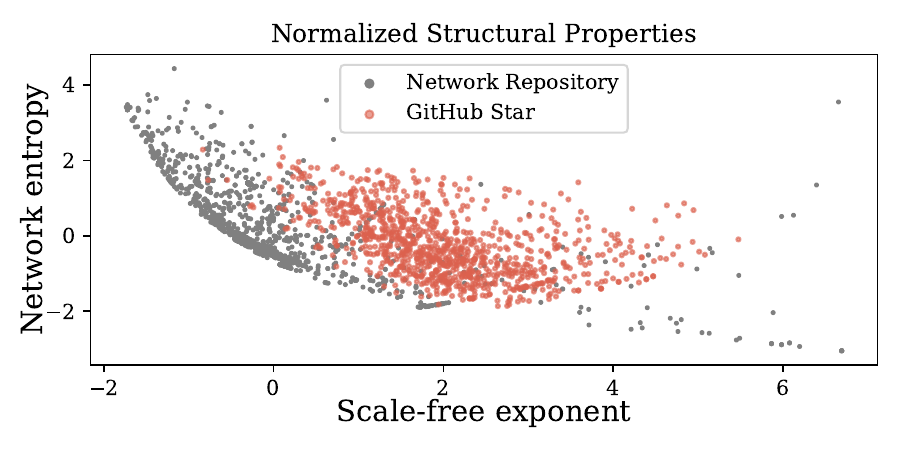}
    \end{center}
    % \vspace{-1.em}
    \caption{
    \small
    Normalized structural properties of Network Repository and Github Star. We enlarge the distribution coverage of our collection by combining both datasets.}
    \label{fig: nr_and_github}
    % \vspace{-0.75em}
% \end{figure}
\end{wrapfigure}

% \jt{I still have concerns about the data collection?? I think it is better to first introduce to achieve our goal 
% "understand the diverse structure patterns of graphs", what are our expectations of the pretraining data and then how  we meet the expections. The followint content is not well motivated like chooing which repository, incompeteness.....  }
% \wenzhuo{motivate: diverse data}
In light of the data scaling spirit, we expect our pre-training data to contain diverse data patterns with sufficient volume.
As graphs from different domains exhibit different patterns~\citep{milo2002network}, we wish to build a collection of graphs from numerous domains to enable a universal graph pattern library with pre-training.
% Various graph databases have become accessible to the public over the past decade, including Network Repository~\citep{rossi2015network}, OGB~\citep{hu2020OGB}, and SNAP~\citep{snapnets}.
Within the publicly available graph databases, Network Repository~\citep{rossi2015network} provides a comprehensive collection of graphs with varied scales from different domains, such as biological networks, chemical networks, social networks, and many more.
Among the thousands of graphs in the Network Repository, some of them contain irregular patterns, including multiple levels of edges, extremely high density, et cetera. 
To ensure the quality of the graphs, we analyze the graph properties following Xu et al.~\citep{xu2023better} and 
% \jt{we still need to manually filter the data so the manually in the intro is not quite good or an advantage of our method???} 
filter out the outliers.
% \jt{if you want to keep this subsection here, you can show "incomplete" here and after you add 1000 graphs, does the incomplete change???}
In addition, we observe that the coverage of graphs in the Network Repository is incomplete according to the network entropy and scale-free exponent, as we observe a relatively scattered space in the middle of Fig.~\ref{fig: nr_and_github}.
% Appendix~\ref{app: datasets}. 
To fill in the gap,  we include a subset of the GitHub Star dataset~\citep{rozemberczki2020karate} by random sampling $1000$ graphs into our graph collection. The selected graphs are utilized to train a discrete diffusion model. 
% which will be detailed in the next section. 

% \begin{figure}[!htp]
%     \centering
%     \vspace{-1em}
%     \resizebox{\textwidth}{!}{
%     \includegraphics{figures/nr_and_github.pdf}
%     }
%     \caption{Illustration of data coverage.}
%     \label{fig: nr_and_github}
%     \vspace{-1em}
% \end{figure}

% Specifically, graphs from different domains exhibit significant variations in scales and semantics, complicating the goal of building a generalized model.

% \jt{instead of focusing on the model, we should focus on the function we want to achieve: for example, we can say: xxxx pretrianing via self-conditioned ...., our downstream adaptation via ....... }

% \subsection{Self-conditioned discrete diffusion model}
% \vspace{-1em}
\subsection{Pre-training through diffusion model}
\label{sec: self-cond}
% \vspace{-0.5em}

% \summary{why diffusion: showcased in CV} 
Diffusion models have demonstrated the ability to facilitate transferability from a data augmentation perspective on the images~\citep{trabucco2024effective, you2024diffusion, he2023is}.
% \mao{add more concrete description on existing success}. 
% \summary{how it works for diffusion model: generating synthetic data} 
Unlike the traditional hand-crafted data augmentation methods, diffusion models can produce more diverse patterns with high quality~\cite{trabucco2024effective}.
With the aid of diffusion guidance~\citep{ho2022classifier, dhariwal2021diffusion}, these methods can achieve domain customization tailored to specific semantic spaces~\citep{you2024diffusion, he2023is}.
% A well-trained diffusion model is utilized to enhance the dataset by generating synthetic data\mao{structure}, where diffusion guidance~\citep{ho2022classifier, dhariwal2021diffusion, gruver2024protein} bridges the gap between pre-trained prior and downstream requirements. 
% \summary{challenges for diffusion models on graphs: 1. non-Euclidean with diverse scale} 
Despite the success of data augmentation through diffusion models on images, the non-Euclidean nature of graph structures poses challenges for data-centric learning on graphs. 
% \summary{challenges for diffusion models on graphs: 2. unlabeled and unfeatured} 
In addition, the fact that most graphs in the Network Repository are unlabeled exacerbates the challenges, as the absence of labeled data results in substantially lower generation quality for diffusion models~\citep{dhariwal2021diffusion, bao2022conditional}.
% conditional diffusion models outperform their unconditional counterparts in terms of both distribution approximation and generation quality~\citep{rombach2022high, zhang2023adding}, which requires .
% In addition, most graphs in the Network Repository are unlabeled, further complicating the goal of building a generalized model.

% An ideal backbone should be both efficient and effective. \mao{To address those challenges} 
To address the aforementioned challenges, we propose to construct a self-conditioned discrete diffusion model on graph structures. 
Unlike Gaussian-based diffusion models, discrete diffusion models~\citep{hoogeboom2021argmax, austin2021structured, campbell2022continuous, vignac2023digress} operate with discrete transition kernels between latent variables, as shown in Section~\ref{sec: background}. 
The key reason we opt for the discrete diffusion models lies in the sparse nature of graphs, where adding Gaussian noise into the adjacency matrix will result in a dense graph~\citep{haefeli2022diffusion}.
On the contrary, discrete diffusion models effectively preserve the sparse structure of graphs during the diffusion process, thus maintaining the efficiency of the models on graphs.
% of message passing in GNNs~\citep{kipf2017semisupervised, hamilton2017inductive}.

% \summary{solve challenge 2: conditional diffusion} 
% Regarding effectiveness, conditional diffusion models outperform their unconditional counterparts in terms of both distribution approximation and generation quality~\citep{rombach2022high, zhang2023adding}. 
% \summary{for unlabeled graph: self-labeling} 
% To improve the performance of diffusion models unlabeled graphs
To accommodate for unlabeled graphs, we adopt a self-supervised labeling strategy as an auxiliary conditioning procedure~\citep{gao2022large, hu2023self}. 
% \summary{why self-labeling helps: better identify diverse graphs} 
By leveraging the self-labeling technique, we are able to upscale the diffusion model to data with more diverse patterns~\citep{gao2022large}. 
The self-labeling technique requires two components: a feature extractor and a self-supervised annotator.
% \summary{why self-labeling helps: better identify diverse graphs}
% \summary{self-labeling helps what: scale up data} Through the self-labeling process, we effectively scale up \method to diverse unlabeled and unfeatured data, thereby providing a powerful prior for graphs from various domains.

% Diffusion models exhibit accurate distribution estimation and high-quality generation when labels are provided on the images~\citep{song2021scorebased, rombach2022high}. While the conditional diffusion models deliver impressive generation quality, they require ground truth annotations for each input sample, which is an unrealistic assumption in the graph domain.
% Specifically, most graphs in the Network Repository do not have ground truth labels. 
% In addition, labels of different graphs typically have distinct semantics. For instance, macromolecules are annotated by catalyzed chemical reactions, while recommendation graphs are labeled by customers' preferences. 

% To eliminate label heterogeneity and provide sample-level conditions, we utilize the self-supervised learning strategy to assign graph-level labels, as inspired by Hu et al.~\citep{hu2023self}. By leveraging the self-labeling process, we are able to upscale the diffusion model to more diverse data~\citep{gao2022large}. The self-labeling process requires two components, i.e., a feature extractor and a self-supervised annotator.

\textbf{Feature extractor.} We extract graph-level features by calculating graph properties, including the number of nodes, density, network entropy, average degree, degree variance, and scale-free exponent following~\cite{xu2023better}.
The first two represent the scale of the graph corresponding to nodes and edges, and the rest indicate the amount of information contained within a graph~\citep{xu2023better}. 
% \mao{In real-world scenarios, the last five characterize different aspects of the graphs, while they are typically highly correlated with each other.} 
We compute the properties of one graph and concatenate them to get a graph-level representation.

\textbf{Self-supervised annotator.} To assign labels to graphs in a self-supervised manner, we employ clustering algorithms on the graph-level representations. 
The number of clusters is determined jointly by the silhouette score~\citep{rousseeuw1987silhouettes} and the separation of the graphs. 
% \mao{Specifically, we visualize the graph properties with clustering labels}. 
The candidates of the number of clusters are chosen to ensure different clusters are well separated. 
Among the candidates, we select the final number of clusters by maximizing the mean Silhouette Coefficient of all samples.

Next we detail the parameterization of the denoising model $\tilde{p}_{\theta}(\widetilde{\mathbf{A}}^{0} \mid \mathbf{A}^{t})$ with the self-assigned graph-level labels $\mathbf{k}$. The denoising model recovers the edges of the original adjacency matrix by predicting the connectivity of the upper triangle, which can be formulated as a link prediction problem~\citep{zhang2018link, kumar2020link}. Following the link prediction setup, the denoising model is composed of a graph transformer (GT)~\citep{shi2020masked} and an MLP link predictor. Denote the hidden dimension as $d$, we treat the node degrees as node features and utilize a linear mapping $f_d: \mathbb{R} \mapsto \mathbb{R}^d$ to match the dimension. Similarly, we utilize another linear mapping $f_t: \mathbb{R} \mapsto \mathbb{R}^d$ for timestep $t$ and learnable embeddings $f_k: \{0, \dots, K\} \mapsto \mathbb{R}^d$ for labels $\mathbf{k}$, where $K$ is the number of clusters. The outputs are summed together and then fed into the GT. Mathematically, we have
% \vspace{-0.5em}
\begin{equation}
\begin{split}
    &\mathbf{h}^t = \operatorname{GT}\left( f_d\left(\operatorname{degree}\left(\mathbf{A}^t\right)\right) + f_t(t) + f_k(\mathbf{k}),\mathbf{A}^t \right), \\
    &\tilde{p}_{\theta}(\widetilde{\mathbf{A}}^{0}_{ij} \mid \mathbf{A}^{t}; t, \mathbf{k}) \coloneq \tilde{p}_{\theta}(\widetilde{\mathbf{A}}^{0}_{ij} \mid \mathbf{h}^t) = \operatorname{MLP} \left( [\mathbf{h}^t_i, \mathbf{h}^t_j] \right).
\end{split}
\end{equation}
With the above denoising network, our diffusion model is trained on the collected graphs by optimizing the variational lower bound in \eqref{eq: vlb}. After the pre-training process, we perform adaptive downstream enhancement through graph structure augmentation.
% which will be illustrated in the following section.

\begin{table*}[!tp]\centering
\caption{
\small
Comparison between GDA methods, pre-training methods, and \method .
By cross-domain transfer, we emphasize the ability of the method to train on vastly different domains and benefit all of them.
% like molecules and social networks, and benefit all domains.
}\label{tab: method_comp}
% \vspace{-0.5em}
\scriptsize
\begin{tabular}{lccccccccccc}\toprule
&\multicolumn{4}{c}{GDA methods} &\multicolumn{3}{c}{Pre-training methods} &\multirow{2}{*}{\method} \\
\cmidrule(lr){2-5} \cmidrule(lr){6-8}
&GraphAug &CFLP &Half-Hop &FLAG &AttrMask &D-SLA &GraphMAE & \\
\midrule
Effective on graph-level task &\textcolor{Green}{\checkmark} &-- &-- &\textcolor{Green}{\checkmark} &\textcolor{Green}{\checkmark} &\textcolor{Green}{\checkmark} &\textcolor{Green}{\checkmark} &\textcolor{Green}{\checkmark} \\
Effective on edge-level task &-- &\textcolor{Green}{\checkmark} &-- &\textcolor{Green}{\checkmark} &-- &\textcolor{Green}{\checkmark} &-- &\textcolor{Green}{\checkmark} \\
Effective on node-level task &-- &-- &\textcolor{Green}{\checkmark} &\textcolor{Green}{\checkmark} &-- &-- &\textcolor{Green}{\checkmark} &\textcolor{Green}{\checkmark} \\
% Handling unlabeled graph &-- &\textcolor{Green}{\checkmark} &\textcolor{Green}{\checkmark} &-- &\textcolor{Green}{\checkmark} &\textcolor{Green}{\checkmark} &\textcolor{Green}{\checkmark} &\textcolor{Green}{\checkmark} \\
% Handling unfeatured graph &-- &-- &\textcolor{Green}{\checkmark} &-- &-- &\textcolor{Green}{\checkmark} &-- &\textcolor{Green}{\checkmark} \\
In-domain transfer &-- &-- &-- &-- &\textcolor{Green}{\checkmark} &\textcolor{Green}{\checkmark} &\textcolor{Green}{\checkmark} &\textcolor{Green}{\checkmark} \\
Cross-domain transfer &-- &-- &-- &-- &-- &-- &-- &\textcolor{Green}{\checkmark} \\
\bottomrule
\end{tabular}
% \vspace{-1em}
\end{table*}

% \vspace{-0.5em}
% \subsection{Structure augmentation through diffusion model}
\subsection{Downstream adaptation through data augmentation}
\label{sec: guidance}
% \vspace{-0.5em}
% \subsection{Guidance strategy across granularity}
% \summary{how to ensure positive transfer: guided generation} Another key component of data augmentation via diffusion models is the guidance strategy. 
% \summary{recall: different tasks have different inductive biases} Downstream tasks on graphs, including node-level, edge-level, and graph-level tasks, require task-specific inductive biases~\citep{luan2024graph, mao2024demystifying, mao2024revisiting, xu2018powerful}. 
% \summary{how to account for different inductive biases: guided generation}To accommodate these diverse tasks, we meticulously design a series of guidance objectives with varying granularity. 
% \summary{how it helps: align generation with downstream} This guidance process serves to bridge the gap between the prior provided by \method and the downstream datasets, incorporating external information to enhance the data. 
% \summary{results in performance boost} Empirically, we apply \method to graphs from various domains and consistently observe performance improvements in node classification, link prediction, and graph classification. 
% To the best of our knowledge, this study represents the first demonstration of the positive transferability of data-centric learning on graphs across domains.

% \summary{how to ensure positive transfer: guided generation} 
% Another key component of data augmentation via diffusion models is the guidance strategy. 

% \summary{how it helps: align generation with downstream} 
The downstream phase of \method is to augment the graph topology through guided generation.
% the downstream dataset by generating synthetic graphs to meet various requirements across different levels of granularity, where the augmented graph is composed of \textit{generated structures} and \textit{original node features}s.
This guidance process serves to provide downstream semantics for the diffusion model, thus bridging the gap between the pre-training distribution and the downstream datasets.
% This guidance process serves to bridge the gap between the prior distribution provided by the pre-trained diffusion model and the downstream datasets, incorporating pre-training knowledge to enhance the data. 
% Diffusion guidance~\citep{ho2022classifier, dhariwal2021diffusion} has greatly improved the sampling quality of diffusion models. 
Among the techniques for diffusion guidance, gradient-based methods~\citep{dhariwal2021diffusion, gruver2024protein} offer versatile approaches by incorporating external conditions that are not present during training. 
For the discrete diffusion process,
% existing methods provide non-gradient guidance~\citep{chen2023dexplainer, vignac2023digress} and gradient guidance~\citep{gruver2024protein} methods. 
we opt for the gradient-based NOS method~\citep{gruver2024protein} due to its flexibility and efficiency. 
Specifically, we build an \textit{MLP regression head} $g_{\theta}: \mathbb{R}^d \mapsto \mathbb{R}^{r}$ that takes the hidden representations $\mathbf{h}^t$ as the input and outputs the guidance objective of dimension $r$.
Denote $\tau$ as the temperature, $\gamma$ as the step-size, $\lambda$ as the regularization strength, and $\varepsilon$ drawn from $\mathcal{N}(0, I)$, we sample from $\tilde{p}^{\prime}(\widetilde{\mathbf{A}}^0 \mid \mathbf{h}^{t}) \propto \tilde{p}_{\theta}(\widetilde{\mathbf{A}}^{0} \mid \mathbf{h}^t) \exp \left(g_{\theta}\left(\mathbf{h}^{t}\right)\right)$ via Langevin dynamics
% \vspace{-1em}
\begin{equation}
\label{eq: guidance}
% \resizebox{\textwidth}{!}{$
    \mathbf{h}^{t, \prime} \leftarrow \mathbf{h}^{t, \prime}-\gamma \nabla_{\mathbf{h}^{t, \prime}}\left[\lambda \mathrm{KL}\left(\tilde{p}^{\prime}\left(\widetilde{\mathbf{A}}^0 \mid \mathbf{h}^{t, \prime}\right) \Vert\ \tilde{p}^{\prime}\left(\widetilde{\mathbf{A}}^0 \mid \mathbf{h}^{t}\right)\right)-g_{\theta}\left(\mathbf{h}^{t, \prime}\right)\right]+\sqrt{2 \gamma \tau} \varepsilon.
% $}
\end{equation}

% The regression head $g_{\theta}$ is trained on the downstream dataset with different objective
One key question to answer is how to choose the proper guidance objectives. 
Our goal is to find numerical characteristics that can best describe the structural properties of a graph.
This includes supervision signal and self-supervised information on the level of node, edge, and graph.
% according to specific downstream tasks, including node classification, link prediction, and graph classification.
% Different tasks may characterize distinct aspects of the graphs, which leads to task-specific model designs.
% In particular, most node classification methods operate based on the principle of homophily, wherein connected nodes tend to have similar characteristics~\citep{khanam2020homophily, mao2024demystifying}.
% Meanwhile, the primary guidance concepts in link prediction can be categorized into local structural proximity, global structural proximity, and feature proximity~\citep{mao2024revisiting}. 
% On the graph level, expressive GNNs follow the Weisfeiler-Lehman test to distinguish graphs from different classes~\citep{xu2018powerful}. 
% To account for the task-specific requirements with varied granularity, we design our supervised and self-supervised guidance objectives attentively on the level of node, edge, and graph.

\textbf{Node level.} Node labels provide the supervision signal for node classification tasks. 
Beyond node labels, node degrees are a fundamental factor in the evolutionary process of a graph~\citep{liu2011controllability}. 
From the perspective of network analysis, centrality measures indicate the importance of nodes from various viewpoints~\citep{borgatti2005centrality}. 
Empirically, we observe that utilizing different node-level heuristics as guidance targets tends to yield similar outcomes. 
Therefore, we focus on node labels and node degrees.

\textbf{Edge level.} Edge-level heuristics can be broadly classified into two categories: local structural heuristics, such as Common Neighbor and Adamic Adar~\citep{adamic2003friends}, and global structural heuristics, such as Katz~\citep{katz1953new} and SimRank~\citep{jeh2002simrank}. 
Similar to node-level heuristics, empirical observations suggest that different edge-level heuristics tend to yield comparable guidance effects. 
In this work, we focus on the Common Neighbors (CN) heuristic due to its efficiency.
Another edge-level guidance objective is to recover the adjacency matrix from the node representations in a link prediction way, similar to how we parameterize the denoising network.
We anticipate that such link prediction objective helps to align the generated graph with the downstream data on the granularity of edges.

\textbf{Graph level.} Graph labels offer the supervision signal for graph classes or regression targets. 
In addition, we incorporate graph-level properties~\citep{xu2023better} as quantitative measures to bridge the gap between the pre-training distribution and the downstream dataset. 
We empirically observe that graph label guidance offers significantly higher performance boosts compared to properties on graph-level tasks. Therefore, we focus on graph labels in our experiments.

We provide our choice of objectives for each task in Appendix~\ref{app: guidance}. We note that all the above objectives are natural choices inspired by heuristics and downstream tasks. There exist many other self-supervised objectives to be explored, such as community-level spectral change~\citep{tan2024community} and motif occurrence prediction~\citep{rong2020self}. We leave the study of objectives as one future work.
With the diffusion guidance, we assemble the augmented graphs with \textit{generated structures} and \textit{original node features}.
The augmented graphs are then fed into downstream-specific GNNs.
% \wenzhuo{describe downstream application with specific GNN; describe the augmented graph: original node features and generated structure}

% \jt{add a subsection to focus on discussing the advantages of our method via design and put Table 1 here }
% \vspace{-0.5em}

\subsection{Comparison to existing methods}
% \vspace{-0.5em}

The data augmentation paradigm of \method allows us to disentangle the upstream and downstream.
We construct a diffusion model as the upstream component to comprehend the structural patterns of graphs across various domains.
In addition, we leverage downstream inductive biases with downstream-specific models in a plug-and-play manner.
% and we utilize the downstream inductive biases with downstream-specific models in a plug-and-playable manner.
This allows \method to facilitate cross-domain transfer, offering a unified method that benefits graphs across different domains for various downstream tasks.
On the contrary, existing GDA methods are typically designed for specific tasks and hard to transfer to unseen patterns.
In the meantime, existing pre-training methods fail to transfer across domains due to heterogeneity in features and structures.
This comparison highlights the success of \method as a data-scaling graph structure augmentor across domains.
We summarize the comparison between methods in Table~\ref{tab: method_comp}.

% \vspace{-1em}

\section{Experiment}
\label{sec: exp}
% \vspace{-0.5em}
In this section, we conduct experiments to validate the effectiveness of \method. 
We first pre-train our discrete diffusion model on thousands of graphs collected from diverse domains.
For each downstream task, we train an \textit{MLP guidance head} with 
% \jt{where is corresponding??}
corresponding objectives on top of the diffusion model.
We then perform structure augmentation using \method and subsequently train a task-specific GNN on augmented data for prediction. 
Through the experiments, we aim to answer the following research questions:
\begin{itemize}
[noitemsep,topsep=0pt,parsep=0pt,partopsep=0pt, leftmargin=*]
    \item RQ1: Can \method benefit graphs from various domains across different downstream tasks?
    \item RQ2: What is the scaling behavior of \method corresponding to data scale and amount of compute?
    \item RQ3: Which components of \method are effective in preventing negative transfer?
\end{itemize}

\begin{table*}[!tp]\centering
\caption{
\small
Mean and standard deviation of accuracy (\%) with $10$-fold cross-validation on graph classification. The best result is \textbf{bold}. The \colorbox{lightgray}{highlighted} results indicate negative transfer for pre-training methods compared to GIN. The last column is the average rank.}\label{tab: graph_class}
% \vspace{-0.5em}
\scriptsize
\resizebox{\textwidth}{!}{
\begin{tabular}{lcccccccccccccccc}\toprule
&DD &Enzymes &Proteins &NCI1 &IMDB-B &IMDB-M &Reddit-B &Reddit-12K &Collab &A.R. \\
\cmidrule{2-10}
GIN &75.81 ± 6.11 &66.00 ± 7.52 &73.32 ± 4.03 &78.30 ± 3.20 &71.10 ± 2.90 &49.07 ± 2.81 &90.85 ± 1.30 &48.63 ± 1.62 &74.54 ± 2.41 &5.56 \\
\midrule
AttrMask &\cellcolor[HTML]{d9d9d9}72.93 ± 3.09 &\cellcolor[HTML]{d9d9d9}23.66 ± 6.09 &\cellcolor[HTML]{d9d9d9}73.10 ± 3.90 &\cellcolor[HTML]{d9d9d9}77.67 ± 2.53 &71.20 ± 2.40 &\cellcolor[HTML]{d9d9d9}48.00 ± 3.14 &\cellcolor[HTML]{d9d9d9}87.50 ± 3.31 &\cellcolor[HTML]{d9d9d9}48.00 ± 1.60 &75.64 ± 1.52 &8.00 \\
CtxtPred &\cellcolor[HTML]{d9d9d9}75.14 ± 2.67 &\cellcolor[HTML]{d9d9d9}21.67 ± 3.87 &\cellcolor[HTML]{d9d9d9}72.21 ± 4.60 &78.99 ± 1.29 &\cellcolor[HTML]{d9d9d9}70.70 ± 1.55 &\cellcolor[HTML]{d9d9d9}48.20 ± 2.23 &\cellcolor[HTML]{d9d9d9}90.35 ± 2.31 &\cellcolor[HTML]{d9d9d9}47.62 ± 2.50 &75.60 ± 1.49 &7.67 \\
EdgePred &\cellcolor[HTML]{d9d9d9}75.64 ± 2.77 &\cellcolor[HTML]{d9d9d9}22.00 ± 3.32 &\cellcolor[HTML]{d9d9d9}71.22 ± 3.53 &\cellcolor[HTML]{d9d9d9}77.82 ± 2.95 &\cellcolor[HTML]{d9d9d9}70.20 ± 2.23 &\cellcolor[HTML]{d9d9d9}47.80 ± 2.42 &\cellcolor[HTML]{d9d9d9}90.80 ± 1.69 &\cellcolor[HTML]{d9d9d9}48.35 ± 1.44 &74.64 ± 2.24 &8.56 \\
InfoMax &\cellcolor[HTML]{d9d9d9}75.23 ± 3.43 &\cellcolor[HTML]{d9d9d9}22.50 ± 6.76 &\cellcolor[HTML]{d9d9d9}71.30 ± 5.18 &\cellcolor[HTML]{d9d9d9}76.94 ± 1.48 &71.60 ± 2.06 &\cellcolor[HTML]{d9d9d9}46.70 ± 2.46 &\cellcolor[HTML]{d9d9d9}89.15 ± 2.84 &48.98 ± 1.83 &75.44 ± 1.12 &8.00 \\
JOAO &75.98 ± 2.86 &\cellcolor[HTML]{d9d9d9}22.17 ± 3.67 &\cellcolor[HTML]{d9d9d9}71.57 ± 5.31 &\cellcolor[HTML]{d9d9d9}76.87 ± 2.27 &\cellcolor[HTML]{d9d9d9}71.02 ± 1.81 &\cellcolor[HTML]{d9d9d9}48.85 ± 2.06 &\cellcolor[HTML]{d9d9d9}90.17 ± 2.13 &49.01 ± 1.90 &74.77 ± 1.71 &7.11 \\
D-SLA &\cellcolor[HTML]{d9d9d9}74.66 ± 3.30 &\cellcolor[HTML]{d9d9d9}22.67 ± 4.21 &\cellcolor[HTML]{d9d9d9}71.97 ± 4.17 &\cellcolor[HTML]{d9d9d9}77.95 ± 2.11 &71.92 ± 2.75 &\cellcolor[HTML]{d9d9d9}47.28 ± 1.88 &\cellcolor[HTML]{d9d9d9}89.77 ± 1.87 &\cellcolor[HTML]{d9d9d9}48.50 ± 1.33 &75.99 ± 2.08 &7.00 \\
GraphMAE &76.07 ± 3.25 &\cellcolor[HTML]{d9d9d9}23.00 ± 3.64 &\cellcolor[HTML]{d9d9d9}70.45 ± 4.19 &79.08 ± 2.72 &71.50 ± 2.01 &\cellcolor[HTML]{d9d9d9}47.93 ± 3.03 &\cellcolor[HTML]{d9d9d9}86.10 ± 3.63 &\cellcolor[HTML]{d9d9d9}47.67 ± 1.16 &74.84 ± 1.36 &7.67 \\
\midrule
S-Mixup &73.12 ± 3.27 &66.85 ± 7.04 &74.61 ± 5.08 &78.91 ± 1.61 &69.61 ± 4.43 &48.33 ± 5.36 &88.65 ± 3.12 &48.30 ± 2.50 &75.89 ± 3.26 &6.67 \\
GraphAug &75.21 ± 2.63 &68.14 ± 7.92 &74.21 ± 3.70 &79.53 ± 3.21 &\textbf{74.00 ± 3.41} &48.11 ± 1.85 &90.50 ± 3.17 &49.00 ± 1.99 &76.02 ± 2.67 &3.67 \\
FLAG &76.87 ± 7.21 &68.35 ± 7.45 &74.31 ± 4.21 &79.03 ± 3.75 &68.83 ± 4.67 &47.21 ± 3.45 &89.11 ± 2.40 &47.48 ± 3.01 &75.32 ± 3.13 &7.00 \\
\midrule
\method &\textbf{78.13 ± 2.61} &\textbf{71.50 ± 5.85} &\textbf{75.47 ± 2.50} &\textbf{80.54 ± 1.77} &73.50 ± 2.48 &\textbf{50.13 ± 2.05} &\textbf{92.28 ± 1.59} &\textbf{49.48 ± 0.71} &\textbf{77.00 ± 2.02} &1.11 \\
\bottomrule
\end{tabular}
}
% \vspace{-1em}
\end{table*}

% \vspace{-0.5em}
\subsection{Main results}
% \vspace{-0.5em}
\label{sec: main_exp}
To get a comprehensive understanding of \method, 
% \jt{I remember in the previous sections, we mention we use thousands of graphs?? please check, it is better to say we use numerous graphs instead of giving a exact number?? }
we evaluate it on $25$ downstream datasets from $7$ domains for graph property prediction, link prediction, and node classification. The statistics of the datasets can be found in Appendix~\ref{app: datasets}, and technical details of the experiments are in Appendix~\ref{app: experiment}.
% Table~\ref{tab: graph_prop_data}, Table~\ref{tab: link_pred_data} and Table~\ref{tab: node_class_data}.

% \textbf{Pre-training data collection} 
% Various graph databases have become accessible to the public over the past decade.
% % including Network Repository~\citep{rossi2015network}, OGB~\citep{hu2020OGB}, and SNAP~\citep{snapnets}.
% Within the publicly available graph databases, Network Repository~\citep{rossi2015network} provides a comprehensive collection of graphs with varied scales from different domains, such as biological networks, chemical networks, social networks, and many more.
% Among the thousands of graphs in the Network Repository, some of them contain irregular patterns, including multiple levels of edges, extremely high density, et cetera. 
% To ensure the quality of the graphs, we analyze the graph properties following Xu et al.~\citep{xu2023better} and conduct graph filtering by excluding the outliers.
% In addition, we observe that the coverage of graphs in the Network Repository is incomplete according to the network entropy and negative scale-free exponent, as illustrated in Fig. 
% \wenzhuo{add a figure} 
% To fill in the gap,  we include a $1000$ graphs subset of the GitHub Star dataset from TUDataset~\citep{morris2020tudataset}. The selected graphs are utilized to train a discrete diffusion model, which will be detailed in the next section.

\paragraph{Baselines.} We evaluate our model against three main groups of baselines. 
(1) Task-specific GNNs: For graph property prediction, we use GIN~\citep{xu2018powerful}; for link prediction, we use GCN~\citep{kipf2017semisupervised} and NCN~\citep{wang2024neural}; and for node classification, we use GCN~\citep{kipf2017semisupervised}. 
(2) Graph pre-training methods: These include AttrMask, CtxtPred, EdgePred, and InfoMax~\citep{hu2019strategies}, JOAO~\citep{you2021graph}, D-SLA~\citep{kim2022graph}, and GraphMAE~\citep{hou2022graphmae}. 
For each of these methods, we pre-train it on the same pre-training set as \method.
While most of the pre-training graphs lack node features, we calculate the node degrees as the input.
Each method consists of three pre-trained variants with different backbone GNNs, including GIN, GCN, and GAT. 
% We note that all these methods require the downstream graphs to have the same node feature space as the pre-training data.
% Therefore, in the fine-tuning stage, we replace the node features of the downstream datasets with node degrees, evaluate all three variants, and report the highest performance for each method in each task.
% We are aware that simply using the node degrees could lead to a decline in performance for the baseline methods.
% Thus, we include more results with \textit{semi-supervised} and \textit{self-supervised} settings in Appendix~\ref{app: experiment}.
(3) Self-supervised methods: These include MVGRL~\citep{hassani2020contrastive}, GRACE~\citep{zhu2020deep}, BGRL~\cite{thakoor2022largescale} and GraphMAE~\citep{hou2022graphmae}.
For the self-supervised methods, we extract the node embeddings and feed them into downstream specific heads.
(4) Graph data augmentation (GDA) methods: For graph property prediction, we include S-Mixup~\citep{ling2023graph}, GraphAug~\citep{luo2022automated}, FLAG~\citep{kong2022robust}, GREA~\citep{liu2022graph}, and DCT~\citep{liu2024data}; for link prediction, we include CFLP~\citep{zhao2022learning}; and for node classification on heterophilic graphs, we include Half-Hop~\citep{azabou2023half}. 
The GDA methods are implemented based on chosen task-specific GNNs.

% \jt{for the experimental settings like data split, and so on, we may need to mention which paper or which standard we follow??}
% and are tuned separately.

% \subsubsection{Graph property prediction}

% \textbf{Experimental settings and implementation details} We collect $12$ datasets from different domains from TUDataset~\citep{morris2020tudataset} and OGB~\citep{hu2020OGB}, including biological networks DD, Enzymes, and Proteins; chemistry network NCI1, ogbg-Lipo, ogbg-ESOL, ogbg-FreeSolv; social networks IMDB-Binary, IMDB-Multi, Reddit-Binary, and Reddit-Multi-12K; and academic network Collab. 
% The statistics of datasets can be found in Appendix~\ref{app: datasets} Table~\ref{tab: graph_prop_data}. We follow the standard $10$-fold cross-validation setting for all the datasets~\citep{Errica2020A}. The results are evaluated by accuracy on the test set. 
% We utilize graph label guidance throughout the graph-level tasks for \method by training a $2$-layer MLP as guidance head on the graph labels in the training set.
% In the augmentation stage, we generate multiple graphs per training sample.
% The generated graphs are then fed into a GIN for graph classification.

% \input{tables/link_pred}

% \input{tables/node_class_hete}

\begin{table*}[!tp]\centering
% \vspace{-1em}
\caption{
\small
Mean and standard deviation across $10$ runs on link prediction. Results are scaled $\times 100$. The last two methods are based on NCN, while the rest are GCN-based. The best result is \textbf{bold} for two backbones, respectively. 
% The \colorbox{lightgray}{highlighted} results indicate negative transfer for pre-training methods compared to GCN. 
The last column is the average rank of each GCN-based method.}\label{tab: link_pred_comp}
% \vspace{-0.5em}
\scriptsize
\resizebox{\textwidth}{!}{
\begin{tabular}{lccccccccc}\toprule
& &Cora &Citeseer &Pubmed &Power &Yeast &Erdos &Flickr &\multirow{2}{*}{A.R.} \\
& &MRR &MRR &MRR &Hits@10 &Hits@10 &Hits@10 &Hits@10 & \\
\cmidrule{2-10}
&GCN &30.26 ± 4.80 &50.57 ± 7.91 &16.38 ± 1.30 &30.61 ± 4.07 &24.71 ± 4.92 &35.71 ± 2.65 &8.10 ± 2.58 &5.14 \\
\midrule
\multirow{4}{*}{Self-supervised} &MVGRL &29.13 ± 3.90 &51.32 ± 4.12 &15.21 ± 2.35 &31.71 ± 3.78 &23.74 ± 5.74 &36.21 ± 2.81 &8.42 ± 2.18 &5.29 \\
&GRACE &31.77 ± 4.31 &49.13 ± 3.95 &16.88 ± 1.74 &28.21 ± 5.04 &23.96 ± 4.31 &33.90 ± 2.12 &\textbf{9.87 ± 0.98} &5.14 \\
&BGRL &33.59 ± 2.14 &51.91 ± 5.01 &16.93 ± 2.03 &33.71 ± 3.21 &25.91 ± 3.12 &37.95 ± 1.73 &8.52 ± 1.85 &3.00 \\
&GraphMAE &32.98 ± 5.01 &52.71 ± 5.39 &\textbf{18.83 ± 1.30} &32.81 ± 2.12 &26.51 ± 2.92 &35.63 ± 3.61 &7.01 ± 3.86 &3.43 \\
\midrule
\multirow{2}{*}{GDA} &CFLP &33.62 ± 6.44 &\textbf{55.20 ± 4.16} &17.01 ± 2.75 &16.02 ± 8.31 &24.23 ± 5.23 &28.74 ± 2.38 &0.00 ± 0.00 &4.57 \\
&UniAug - GCN &\textbf{35.36 ± 7.88} &54.66 ± 4.55 &17.28 ± 1.89 &\textbf{34.36 ± 1.68} &\textbf{27.52 ± 4.80} &\textbf{39.67 ± 4.51} &9.46 ± 1.18 &1.43 \\
\midrule
\multirow{2}{*}{NCN-based} &NCN &31.72 ± 4.48 &58.03 ± 3.45 &38.26 ± 2.56 &27.36 ± 5.00 &39.85 ± 5.07 &36.81 ± 3.29 &8.33 ± 0.92 &-- \\
&UniAug - NCN &\textbf{35.92 ± 7.85} &\textbf{61.69 ± 3.21} &\textbf{40.30 ± 2.53} &\textbf{30.20 ± 1.46} &\textbf{42.11 ± 5.74} &\textbf{39.26 ± 2.84} &\textbf{8.85 ± 0.90} &-- \\
\bottomrule
\end{tabular}
}
% \vspace{-1em}
\end{table*}

\begin{table*}[!tp]\centering
% \caption{Generated by Spread-LaTeX}\label{tab: }
\caption{
\small
Mean and standard deviation of accuracy (\%) across $10$ splits on node classification of heterophilic graphs. The best result is \textbf{bold}. 
% The \colorbox{lightgray}{highlighted} results indicate negative transfer for pre-training methods compared to GCN. 
The last column is the average rank of each method.}\label{tab: node_class_hete_comp}
% \vspace{-0.5em}
\scriptsize
\resizebox{\textwidth}{!}{
\begin{tabular}{lcccccccccccc}\toprule
& &Cornell &Wisconsin &Texas &Actor &Chameleon* &Squirrel* &A.R. \\
\cmidrule{2-9}
&GCN &59.41 ± 6.03 &51.68 ± 4.34 &63.78 ± 4.80 &30.58 ± 1.29 &40.94 ± 3.91 &39.11 ± 1.74 &4.50 \\
\midrule
\multirow{4}{*}{Self-supervised} &MVGRL &56.19 ± 2.42 &50.64 ± 5.89 &61.70 ± 3.94 &31.37 ± 0.83 &32.34 ± 2.11 &35.32 ± 1.32 &6.83 \\
&GRACE &56.39 ± 2.11 &53.83 ± 3.56 &63.54 ± 2.57 &28.14 ± 0.81 &35.71 ± 1.95 &33.65 ± 2.51 &7.00 \\
&BGRL &56.67 ± 2.13 &59.80 ± 4.08 &65.78 ± 2.66 &29.80 ± 0.31 &37.01 ± 2.89 &34.77 ± 2.01 &5.33 \\
&GraphMAE &57.31 ± 2.11 &58.27 ± 2.91 &58.34 ± 3.57 &28.97 ± 0.27 &36.75 ± 1.78 &39.13 ± 2.01 &5.50 \\
\midrule
\multirow{3}{*}{GDA} &Half-Hop &62.46 ± 7.58 &76.47 ± 2.61 &72.35 ± 4.27 &33.95 ± 0.68 &38.59 ± 2.89 &37.34 ± 2.18 &3.17 \\
&UniAug &68.11 ± 6.72 &69.02 ± 4.96 &73.51 ± 5.06 &33.11 ± 1.57 &\textbf{43.84 ± 3.39} &\textbf{41.90 ± 1.90} &2.00 \\
&UniAug + Half-Hop &\textbf{72.43 ± 5.81} &\textbf{79.61 ± 5.56} &\textbf{77.03 ± 4.27} &\textbf{34.97 ± 0.55} &41.94 ± 2.77 &38.79 ± 2.61 &1.67 \\
\bottomrule
\end{tabular}
}
% \vspace{-1em}
\end{table*}

\begin{wraptable}{r}{0.55\textwidth}
% \begin{table}[!htp]
\vspace{-1em}
\begin{threeparttable}
\caption{\small Mean and standard deviation of MAE $\downarrow$ across $10$ runs on molecule regression. The last column is the average rank of each method. Among the methods, all pre-training methods discard atom and bond features due to dimension mismatch and we include the best-performing method JOAO into comparison; GIN and \method remove the bond features; others incorporate both. }\label{tab: graph_reg}
% \vspace{-0.5em}
\scriptsize
\begin{tabular}{lccccc}\toprule
&ogbg-Lipo &ogbg-ESOL &ogbg-FreeSolv &A.R. \\
\cmidrule{2-4}
GINE* &0.545 ± 0.019 &0.766 ± 0.016 &1.639 ± 0.146 &5.00 \\
GIN &0.543 ± 0.021 &0.729 ± 0.018 &1.613 ± 0.155 &3.67 \\
\midrule
JOAO &0.859 ± 0.007 &1.458 ± 0.040 &3.292 ± 0.117 &7.00 \\
\midrule
FLAG* &0.528 ± 0.012 &0.755 ± 0.039 &1.565 ± 0.098 &3.00 \\
GREA* &0.586 ± 0.036 &0.805 ± 0.135 &1.829 ± 0.368 &6.00 \\
DCT* &0.516 ± 0.071 &0.717 ± 0.020 &1.339 ± 0.075 &1.33 \\
\midrule
\method &0.528 ± 0.006 &0.677 ± 0.026 &1.448 ± 0.049 &1.67 \\
\bottomrule
\end{tabular}
\begin{tablenotes}
  \small
  \item *Results are taken from DCT~\citep{liu2024data}.
\end{tablenotes}
\end{threeparttable}
% \vspace{-0.5em}
% \end{table}
\end{wraptable}

\paragraph{Graph property prediction.} 
We employ graph label guidance for \method throughout the graph-level tasks by training a $2$-layer MLP as the guidance head on the graph labels in the training set.
In the augmentation stage, we generate multiple graphs per training sample, and the generated graphs are then fed into the baseline GIN.
We present the results of molecule regression in Table~\ref{tab: graph_reg} and graph classification in Table~\ref{tab: graph_class}. 
Three key observations emerge from the analysis:
(1) Existing pre-training methods show negative transfer compared to GIN. 
Some special cases are the Enzymes and molecule regression datasets, where all pre-training methods fail to yield satisfactory results. 
In these datasets, the features are one of the driving components for graph property prediction, while the pre-training methods fail to encode such information due to incompatibility with the feature dimension.
This reveals one critical drawback of the pre-training methods: their inability to handle feature heterogeneity.
(2) GDA methods yield inconsistent results across different datasets. 
While these methods enhance performance in some datasets, they cause performance declines in others. 
This variability is directly reflected in the average rank, where some of them even fall behind the GIN.
(3) Unlike the pre-training methods and GDA methods, \method shows consistent performance improvements against GIN with a large margin.
In the molecule regression tasks, \method effectively compensates for the absence of bond features and achieves performance comparable to DCT, which is a data augmentation method pre-trained on in-domain molecule graphs.
% Note that we replace the original node features with node degrees when pre-training the baselines on our graph collection due to missing features and mismatched semantics.
% We understand that removing the node features may result in a performance drop for the baseline methods. 
We also adapt the \textit{semi-supervised}~\citep{you2020l2} and \textit{self-supervised}~\citep{Sun2020InfoGraph} setting for the baselines for a comprehensive benchmark in Appendix~\ref{app: exp_graph_prop_pred} Table~\ref{tab: graph_class_comparison}, where we observe that UniAug presents consistently satisfactory performance according to the average rank, matching or outperforming the best baseline.
These findings affirm that the pre-training and structure augmentation paradigm of \method effectively benefits the downstream datasets at the graph level.

\paragraph{Link prediction.}
We choose three guidance objectives for \method, including node degree, CN, and link prediction objective, as described in Section~\ref{sec: guidance}.
For each objective, we train an MLP to provide guidance information.
We then augment the graph structure by generating a synthetic graph and preserving the original training edges, ensuring that the augmented graph does not remove any existing edges.
The augmented graph is then fed into a GCN for link prediction.
We summarize the results in Table~\ref{tab: link_pred_comp}, where we observe that
% (1) Existing pre-training methods provide negative transfer, especially on datasets with node features. 
(1) GDA method CFLP leads to performance drops on the datasets without features and also suffers from high computation complexity during preprocessing. 
(2) \method enhances performance across all tested datasets.
In addition, we employ \method to NCN~\citep{wang2024neural}, one of the state-of-the-art methods for link prediction.
% The results, \jt{why not merge table 4 and 6 for link prediction??? you can denote GCN as basebone as UniAug-GCN, similarly for NCN??} detailed in Table~\ref{tab: link_pred_ncn}, 
The results demonstrate consistent performance boosts from \method when we apply NCN as the backbone.
The structure augmentation paradigm of \method allows plug-and-play applications to any downstream-specific models, showcasing its adaptability and effectiveness. 
Additional results, including the performance of pre-training baselines and the effects of three guidance objectives, can be found in Appendix~\ref{app: link_pred}.
% In addition, we study the effects of three guidance objectives. More details can be found in Appendix~\ref{app: link_pred}.

% \begin{table}[!tp]\centering
\begin{wraptable}{r}{0.5\textwidth}
\vspace{-1em}
\caption{
\small
Results of node classification on homophily graphs. Results are scaled $\times 100$.
}
\label{tab: node_class_homo}
% \vspace{1em}
% \vspace{-0.5em}
\scriptsize
\resizebox{0.5\textwidth}{!}{
\begin{tabular}{lcccccc}\toprule
& &Cora &Citeseer &Pubmed \\
\midrule
\multirow{2}{*}[-0.em]{ACC $\uparrow$} &GCN &81.75 ± 0.73 &70.71 ± 0.76 &79.53 ± 0.25 \\
% \cmidrule(lr){2-5}
&\method &81.78 ± 0.60 &71.17 ± 0.58 &79.54 ± 0.35 \\
% \cmidrule(lr){0-1} \cmidrule(lr){3-5}
\midrule
\multirow{2}{*}[-0.em]{SD $\downarrow$} &GCN &24.51 ± 1.06 &22.57 ± 0.80 &27.02 ± 0.56 \\
% \cmidrule(lr){2-5}
&\method &\textbf{23.45 ± 0.90} &\textbf{19.90 ± 0.81} &\textbf{26.50 ± 0.55} \\
\bottomrule
\end{tabular}
}
% \vspace{-1em}
\end{wraptable}
% \end{table}

\paragraph{Node classification.}
To demonstrate the effectiveness of \method in node-level tasks, we transform the node classification into subgraph classification.
Specifically, we extract the aggregation tree of each node, i.e., $2$-hop subgraph for a $2$-layer GCN, and label the subgraph with the center node.
We then adopt a strategy similar to graph classification and train a $2$-layer classifier as a guidance head.
% \jt{I think we should clarify why here we focus on heterophilic graphs?? I think people may question this set of experiments. let us be careful here }
Inspired by the success of structure augmentation on heterophilic graphs~\citep{bi2022make, azabou2023half}, we evaluate \method on $6$ heterophilic datasets.
We observe phenomena similar to those seen in graph- and link-level tasks in Table~\ref{tab: node_class_hete_comp}, with results of pre-training baselines in Appendix~\ref{app: node_class}.
One thing to mention is the combination of \method and Half-Hop.
Half-Hop offers performance improvements in four out of six datasets via data augmentation, and combining it with \method yields even higher results.
This highlights the flexibility of \method and opens up possibilities for further exploration of its use cases. 
Given the impressive results of \method on heterophilic graphs, we anticipate it will also help to balance the performance disparities among nodes with different homophily ratios on homophilic graphs~\citep{mao2024demystifying}.
We split the nodes into five groups according to their homophily ratios and calculate the standard deviation (SD) across groups.
As shown in Table~\ref{tab: node_class_homo}, \method matches the performance of vanilla GCN and also reduces the performance discrepancies corresponding to SD.

% \input{tables/link_pred_ncn}

% \input{tables/node_class_hete}

% \vspace{-0.5em}
\subsection{Scaling behavior of \method}
% \vspace{-0.5em}
\label{sec: scaling}
% \wenzhuo{may try to draw figures instead of tables}

% In this section, we investigate the scaling behavior of \method in terms of data scale and pre-training time. 

In light of the neural scaling law~\citep{kaplan2020scaling, hoffmann2022training, abnar2022exploring, zhai2022scaling, liu2024neural},
% on natural language~\citep{kaplan2020scaling, hoffmann2022training}, images~\citep{abnar2022exploring, zhai2022scaling}, and graphs~\citep{liu2024neural}, 
we expect \method to benefit from an increased coverage of data and more compute budget.
% In addition, we aim to explore the pre-training process of \method to study the effects of computing power for \method.
In this subsection, we investigate the scaling behavior of \method in terms of data scale and amount of compute for pre-training.

\begin{figure*}[!tp]
    \centering
    % \vspace{-0.5em}
    \begin{subfigure}[t]{0.42857\textwidth}  % 0.42857
        \centering
        % \resizebox{\textwidth}{!}{
        \includegraphics[width=\textwidth, left]{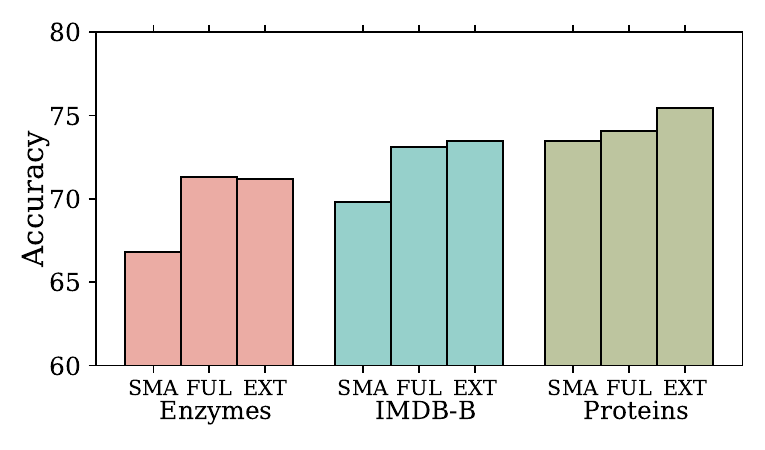}
        % \caption{Lorem ipsum}
    \end{subfigure}%
    ~ 
    \begin{subfigure}[t]{0.57143\textwidth}  % 0.57143
        \centering
        \includegraphics[width=\textwidth, right]{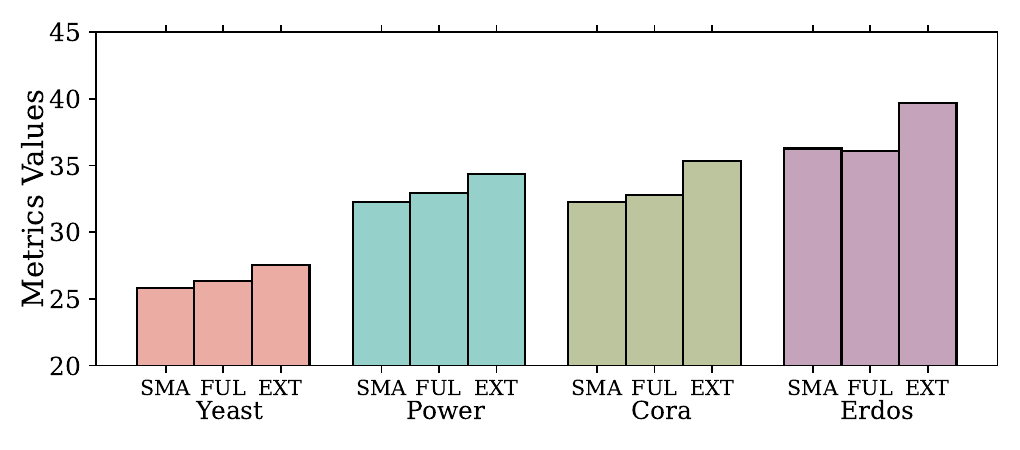}
        % \caption{Lorem ipsum, lorem ipsum,Lorem ipsum, lorem ipsum,Lorem ipsum}
    \end{subfigure}
    \vspace{-1em}
    \caption{
    \small
    Effects of pre-training data scale on graph classification (left) and link prediction (right). The groups SMA, FUL, and EXT represent SMALL, FULL, and EXTRA data collection.
    }
    \label{fig: data}
    % \vspace{-1em}
\end{figure*}

\paragraph{Data coverage }
During the data collection process, we prepare three versions of the training data with increasing magnitude and growing coverage on the graph distribution.
We first sample $10$ graphs per category from the Network Repository~\citep{rossi2015network} to build a SMALL collection.
Next, we gather all the graphs from the Network Repository and filter out large-scale graphs and outliers for a FULL collection.
In addition, we add a $1000$ graphs subset of the GitHub Star dataset from TUDataset~\citep{morris2020tudataset} to enlarge the coverage of diverse patterns and form an EXTRA collection.
We pre-train three versions of \method respectively on the three collections and evaluate them on graph classification and link prediction.
As shown in Fig.~\ref{fig: data}, we observe a clear trend of increase in performance as we enlarge the coverage of pre-training data.
This paves the way to scale up \method to even more pre-training datasets with an expanding distribution of graphs.
Additional results on scaling effects with respect to the proportion of the full data can be found in Appendix~\ref{app: scaling}.

\begin{figure*}[!tp]
    \centering
    % \vspace{-1.em}
    \begin{subfigure}[t]{0.5\textwidth}
        \centering
        % \resizebox{\textwidth}{!}{
        \includegraphics[width=\textwidth, left]{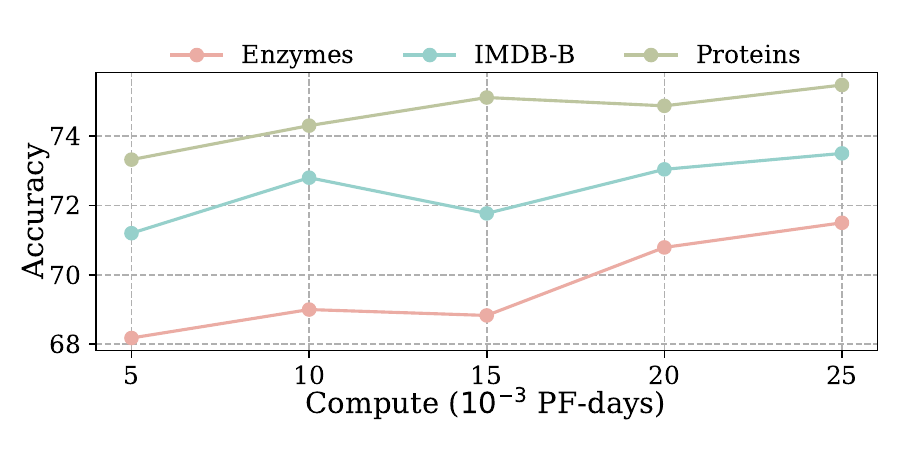}
        % \caption{Lorem ipsum}
    \end{subfigure}%
    ~ 
    \begin{subfigure}[t]{0.5\textwidth}
        \centering
        \includegraphics[width=\textwidth, right]{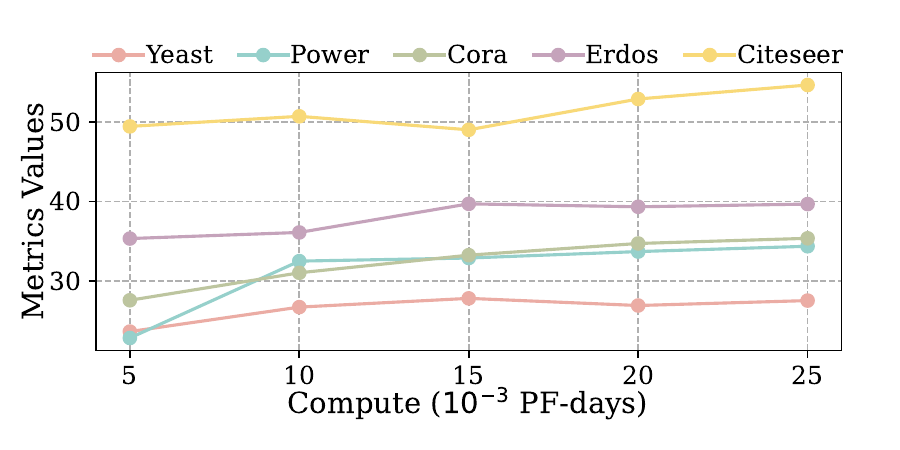}
        % \caption{Lorem ipsum, lorem ipsum,Lorem ipsum, lorem ipsum,Lorem ipsum}
    \end{subfigure}
    \vspace{-1em}
    \caption{
    \small
    Effects of pre-training amount of compute on graph classification (left) and link prediction (right), where one $\text{PF-days} = 10^{15} \times 24 \times 3600 = 8.64 \times 10^{19}$ floating point operations.
    }
    \label{fig: compute}
    % \vspace{-1em}
\end{figure*}

\paragraph{Amount of compute }
We sought to understand how effectively our diffusion model can learn data patterns as we continue to train it.
To this end, we checkpointed \method every 2,000 epochs ($5\times 10^{-3}$ PF-days) while training on the EXTRA collection, and then applied it to graph classification and link prediction tasks.
The results are illustrated in Fig.~\ref{fig: compute}.
We observe that downstream performance generally improves with prolonged training, while the trend slows down for some datasets when we reach 8,000 epochs.
We take the checkpoint at the 10,000th epoch for evaluations.
Given the scaling behavior observed, we anticipate \method to become even more effective with additional resources.

% \vspace{-1em}

% \input{tables/training_time}

\subsection{Preventing negative transfer}

\vspace{-0.5em}

\begin{wraptable}{r}{0.6\textwidth}
% \begin{table}[!tp]
    \vspace{-1em}
    \scriptsize
    \caption{
    \small
    Demonstration of negative transfer on graph classification (up) and link prediction (down).
    }
    % \vspace{-1em}
    \label{tab: ablation}

    \resizebox{0.6\textwidth}{!}{
    \begin{tabular}{lcccccc}\toprule
    &Enzymes &Proteins &IMDB-B &IMDB-M \\
    \midrule
    % \cmidrule(lr){2-5}
    GIN &66.00 ± 7.52 &73.32 ± 4.03 &71.10 ± 2.90 &49.07 ± 2.81 \\
    \midrule
    \method &71.50 ± 5.85 &75.47 ± 2.50 &73.50 ± 2.48 &50.13 ± 2.05 \\
    w/o self-cond &71.11 ± 7.50 &73.31 ± 4.63 &71.50 ± 2.27 &49.00 ± 2.74 \\
    w/o guidance &62.17 ± 3.93 &71.15 ± 4.56 &53.80 ± 3.29 &35.33 ± 3.17 \\
    w/ cross-guide &51.50 ± 7.64 &72.46 ± 4.35 &71.10 ± 2.38 &49.20 ± 2.59 \\
    \bottomrule
    \end{tabular}
    }

    \vspace{1em}
    \resizebox{0.6\textwidth}{!}{
    \begin{tabular}{lcccccc}\toprule
    &Cora &Citeseer &Power &Yeast &Erdos \\
    &MRR &MRR &Hits@10 &Hits@10 &Hits@10 \\
    \midrule
    % \cmidrule(lr){2-6}
    GCN &30.26 ± 4.80 &50.57 ± 7.91 &30.61 ± 4.07 &24.71 ± 4.92 &35.71 ± 2.65 \\
    \midrule
    \method &35.36 ± 7.88 &54.66 ± 4.55 &34.36 ± 1.68 &27.52 ± 4.80 &39.67 ± 4.51 \\
    w/o self-cond &27.97 ± 16.11 &37.65 ± 6.00 &28.95 ± 7.73 &23.54 ± 8.28 &34.33 ± 6.18 \\
    w/o guidance &29.60 ± 6.06 &51.41 ± 7.10 &25.57 ± 6.04 &25.26 ± 6.06 &37.11 ± 4.16 \\
    w/ cross-guide &32.37 ± 4.20 &50.59 ± 5.67 &32.99 ± 2.54 &26.76 ± 3.88 &36.30 ± 3.67 \\
    \bottomrule
    \end{tabular}
    }
    \vspace{-1.5em}
% \end{table}
\end{wraptable}

In the previous parts of the experiments, we showcase the positive transfer of \method across different tasks.
We now investigate which aspects of the design prevent negative transfer. 
\method consists of two main components: a pre-trained diffusion model and the structure augmentation through guided generation.
In the pre-training process, we inject self-supervised graph labels into the diffusion model and we wonder about the performance of its unconditioned counterpart.
Regarding the augmentation process, we examine the impact of diffusion guidance by exploring outcomes when the guidance is either removed or applied using another dataset from a different domain (cross-guide). 
We summarize the results in Table~\ref{tab: ablation} for graph classification and link prediction.
All modifications investigated lead to performance declines in both tasks.
We observe that removing guidance results in significant negative transfers for graph classification, while the effects of self-conditioning are more pronounced for link prediction.
We conclude that both the self-conditioning strategy and diffusion guidance are crucial in preventing negative transfer, underscoring their importance in the design of \method.

\vspace{-0.5em}

\section{Conclusion and Discussion}
\label{sec: conclusion}

In this work, we propose a graph structure augmentor \method to leverage the increasing scale of graph data.
We collect thousands of graphs from various domains and pre-train a self-conditioned discrete diffusion model on them.
In the downstream stage, we augment the graphs by preserving the original node features and generating synthetic structures.
We apply \method to node-, link-, and graph-level tasks and achieve consistent performance gain.
We have successfully developed a showcase that benefits from cross-domain graph data scaling using diffusion models.

% One future direction of the proposed pipeline lies in the parameterization of the diffusion model.
% Our parameterization requires predicting the upper triangle of the adjacency matrix, which results in high spatial complexity for graphs with a large number of nodes.
% We anticipate \emthod to benefit from more efficient 
% A more efficient parameterization will help \method to scale to large graphs.
One limitation of the current analysis is the absence of an investigation into the effects of model parameters due to limited resources.
Given the scaling behavior of \method in terms of data scale and amount of compute, we anticipate that a large-scale model will provide significant performance improvements.
One future direction is to investigate the adaptation of fast sampling methods to the discrete diffusion models on graphs.
This will lead to lower time complexity and enable broader application scenarios.
% Consequently, we have to exclude large graphs from the pre-training set due to limited resources.
% \wenzhuo{no model scale due to limited resources}

\vspace{-0.5em}

\section{Impact Statements}
\label{app: impact}
In this work, we build a universal graph structure augmentor that benefits from data scaling across domains.
Given the consistent performance improvements for different tasks, we expect this work to contribute significantly towards the goal of building a graph foundation model.
In the meantime, we showcase the power of the deep generative models on graphs by introducing new application scenarios.
We anticipate such success will contribute to the community of generative models and graph learning.

It is important to mention that the model backbones of our method and baselines heavily rely on neighboring node information as an inductive bias. 
However, this characteristic can result in biased predictions, especially when patterns in neighborhood majorities dominate, leading to potential ethical issues in model predictions.

\section{Acknowledgment}
W.T. and J.T. are supported by the National Science Foundation (NSF) under grant numbers CNS2321416, IIS2212032, IIS2212144, IIS 2504089, DUE2234015, CNS2246050, DRL2405483 and IOS2035472, the Michigan Department of Agriculture and Rural Development, US Dept of Commerce, Amazon Faculty Award, Meta, NVIDIA, Microsoft and SNAP. W.T. and Y.X. are supported by the National Institutes of Health (NIH) under grant numbers U01DE033330, R01HL166508, R01DE026728 and the NSF under grant numbers OISE2434687 and IOS2107215.

\section{Disclaimer}
 
This paper was prepared for informational purposes by the Artificial Intelligence Research group of JPMorgan Chase \& Co. and its affiliates (``JP Morgan’’), and is not a product of the Research Department of JP Morgan. JP Morgan makes no representation and warranty whatsoever and disclaims all liability, for the completeness, accuracy or reliability of the information contained herein. This document is not intended as investment research or investment advice, or a recommendation, offer or solicitation for the purchase or sale of any security, financial instrument, financial product or service, or to be used in any way for evaluating the merits of participating in any transaction, and shall not constitute a solicitation under any jurisdiction or to any person, if such solicitation under such jurisdiction or to such person would be unlawful.

\bibliographystyle{unsrt}
\bibliography{ref}

%%%%%%%%%%%%%%%%%%%%%%%%%%%%%%%%%%%%%%%%%%%%%%%%%%%%%%%%%%%%

\appendix

\newpage
\appendix
\onecolumn

\section{Derivation of Diffusion Process}
\label{app: diffusion}

In the following, we will formulate the existing discrete diffusion models into binary diffusion on the adjacency matrix. We denote the adjacency matrix of a graph as $\mathbf{A}^0 \in \{0, 1\}^{n\times n}$ with $n$ nodes.
% and denote one element of $\mathbf{A}^0$ as $\mathbf{A}^0$. 
Following D3PM~\citep{austin2021structured}, we corrupt the adjacency matrix into a sequence of latent variables $\mathbf{A}^{1:T} = \mathbf{A}^1, \mathbf{A}^2, \dots, \mathbf{A}^T$ by independently injecting noise into each element with a Markov process
\begin{equation}
    q\left(\mathbf{A}^t \mid \mathbf{A}^{t-1}\right)=\prod_{i,j: i<j} \operatorname{Cat}\left(\mathbf{A}^t_{ij} ; \mathbf{p}=\mathbf{A}^{t-1}_{ij} \mathbf{Q}^t \right),
\end{equation}
where $\mathbf{Q}^t \in [0, 1]^{2\times 2}$ is the transition probability of timestep $t$. The above Markov process is called \textit{forward process}. 
Existing works provide different designs for the transition matrix $\mathbf{Q}^t$, including
\begin{equation}
\resizebox{0.48\textwidth}{!}{$
\begin{split}
    \text{Uniform~\citep{chen2023dexplainer}}:
    &\begin{pmatrix}
        1 - \beta^t & \beta^t \\
        \beta^t & 1 - \beta^t
    \end{pmatrix}; \\
    \hspace{0.5em}
    \text{Absorbing~\citep{chen2023efficient}}:
    &\begin{pmatrix}
        1 & 0 \\
        \beta^t & 1 - \beta^t
    \end{pmatrix}; \\
    \hspace{0.5em}
    \text{Predefined~\citep{vignac2023digress}}:
    &\begin{pmatrix}
        1 - \beta^t \cdot \pi & \beta^t \cdot p \\
        (1-\pi)\beta^t & 1 - (1-\pi)\beta^t
    \end{pmatrix},
\end{split}
$}
% \resizebox{\textwidth}{!}{$
%     \text{Uniform~\citep{chen2023dexplainer}}:
%     \begin{pmatrix}
%         1 - \beta^t & \beta^t \\
%         \beta^t & 1 - \beta^t
%     \end{pmatrix};
%     \hspace{0.5em}
%     \text{Absorbing~\citep{chen2023efficient}}:
%     \begin{pmatrix}
%         1 & 0 \\
%         \beta^t & 1 - \beta^t
%     \end{pmatrix};
%     \hspace{0.5em}
%     \text{Predefined~\citep{vignac2023digress}}:
%     \begin{pmatrix}
%         1 - \beta^t \cdot \pi & \beta^t \cdot p \\
%         (1-\pi)\beta^t & 1 - (1-\pi)\beta^t
%     \end{pmatrix},
% $}
\end{equation}
where $\pi$ is the converging non-zero probability and $\beta^t$ is the noise scale.
All three transition matrices can be written as binary diffusion with Bernoulli distribution
\begin{equation}
\resizebox{0.48\textwidth}{!}{$
\begin{split}
    q\left(\mathbf{A}^t \mid \mathbf{A}^{t-1}\right) &=\operatorname{Bernoulli}\left(\mathbf{A}^t ; \alpha^t \mathbf{A}^{t-1} + \left(1-\alpha^{t}\right) \pi \right), \\
    q\left(\mathbf{A}^t \mid \mathbf{A}^{0}\right) & =\operatorname{Bernoulli}\left(\mathbf{A}^t  ; \bar{\alpha}^{t} \mathbf{A}^0 +\left(1-\bar{\alpha}^{t}\right) \pi \right), \\
    q\left(\mathbf{A}^{t-1} \mid \mathbf{A}^{t}, \mathbf{A}^{0}\right) &= \frac{q\left(\mathbf{A}^{t} \mid \mathbf{A}^{t-1}\right) q\left(\mathbf{A}^{t-1} \mid \mathbf{A}^{0}\right)}{q\left(\mathbf{A}^{t} \mid \mathbf{A}^{0}\right)},
\end{split}
$}
\end{equation}
where $\alpha^t = 1 - \beta^t$ and $\bar{\alpha}^{t} = \prod_{i=1}^t \alpha^i$. 
The prior $\mathbf{A}^T$ is determined by $\pi$ with $p\left(\mathbf{A}^T_{ij}\right) = \operatorname{Bernoulli}(\pi)$, i.e., the existence of each edge follows a Bernoulli distribution with probability $\pi$.
The main difference of the \textit{forward process} among the existing works is the choice of $\pi$, where $\pi=0$ for EDGE~\citep{chen2023efficient}, $\pi=0.5$ for D4Explainer~\citep{chen2023dexplainer}, and a pre-computed average density $\pi$ for DiGress~\citep{vignac2023digress}. 

In our early experiments, we observe that the absorbing kernel $\pi=0$ surpasses the other two in terms of efficiency and effectiveness for graph generation.
The \textit{forward process} with non-zero $\pi$ will add non-existing edges, which brings in additional computations.
When sampling from prior, non-zero $\pi$ will introduce additional uncertainty because we will first sample every edge from $\operatorname{Bernoulli}(\pi)$.
Therefore, we choose the absorbing prior $\pi=0$ in this work and leave the exploration of other transition kernels as a future work.

We note that in our implementation, we choose the number of timesteps $T$ as $128$ according to our early experiments and some existing works~\citep{wang2023binary, chen2023efficient}.
We leave the study of the effects of diffusion timesteps on downstream tasks as a future work.

\section{Guidance Objective for Downstream Tasks}
\label{app: guidance}

We mention various guidance objectives in Section~\ref{sec: guidance} with different granularity.
Here, we specify the objectives we use for each downstream task.
Our empirical results suggest that supervision signals will lead to better performance.
Thus, we use node labels for node classification and graph labels for graph property prediction in Section~\ref{sec: exp}.
Regarding link prediction, we anticipate that both node-level and edge-level objectives may help the downstream adaptation.
Therefore, we choose three objectives including node degree, CN heuristic, and link prediction objective.

\section{Datasets}
\label{app: datasets}

The license of the datasets use in this work is in Table~\ref{tab: license}.
% \subsection{Pre-training data collection}
% As mentioned in Section~\ref{sec: pre_data}, we observe that the coverage of graphs in the Network Repository is incomplete according to the network entropy and scale-free exponent.
% We visualize the coverage in Fig.~\ref{fig: nr_and_github}, where we can see that there is a relatively scattered space in the middle of the figure if we only consider the graphs from the Network Repository.
% We observe that the GitHub Star dataset can fill in the gap, thus we sample a $1000$ graphs subset to enlarge our pre-training data collection.
% As we observe the data scaling behavior in Section~\ref{sec: scaling} for \method, we anticipate that incorporating more data will lead to a more general model.

\begin{table}[!tp]\centering
% \vspace{-2em}
\caption{List of datasets and corresponding License}\label{tab: license}
\scriptsize
% \resizebox{\textwidth}{!}{
% \begin{tabular}{@{}lccccccccccccccccccc@{}}
% \toprule
%  & Network Repository & Github Star & Cora & Citeseer & Pubmed & WebKB & Wikipedia Network & Actor & Power & Yeast & Erdos & Amazon Photo & Flickr & DD & Enzymes & Proteins & NCI1 & IMDB & Reddit \\ \midrule
% License & CC BY-SA & CC BY 4.0 & NLM license & NLM license & NLM license & MIT license & MIT license & MIT license & BSD License & BSD License & BSD License & MIT license & MIT license & CC BY 4.0 & CC BY 4.0 & CC BY 4.0 & CC BY 4.0 & CC BY 4.0 \\ \bottomrule
% \end{tabular}
% }
\begin{tabular}{lcc}\toprule
Dataset &License \\
\midrule
Network Repository &CC BY-SA \\
Github Star &CC BY 4.0  \\
Cora &NLM license \\
Citeseer &NLM license \\
Pubmed &NLM license \\
WebKB &MIT license \\
Wikipedia Network &MIT license \\
Actor &MIT license \\
Power &BSD License \\
Yeast &BSD License \\
Erdos &BSD License \\
Amazon Photo &MIT license \\
Flickr &MIT license \\
DD &CC BY 4.0  \\
Enzymes &CC BY 4.0  \\
Proteins &CC BY 4.0  \\
NCI1 &CC BY 4.0  \\
IMDB &CC BY 4.0  \\
Reddit &CC BY 4.0  \\
\bottomrule
\end{tabular}
% \vspace{-2em}
\end{table}

% \subsection{Downstream datasets}

\textbf{Graph property prediction datasets} include DD and Proteins~\citep{dobson2003distinguishing}, Enzymes~\citep{schomburg2004brenda}, NCI1~\citep{wale2008comparison}, IMDB-Binary, IMDB-Multi, Reddit-Binary, and Reddit-Multi-12K~\citep{yanardag2015deep}, ogbg-Lipo, ogbg-ESOL and ogbg-FreeSolv~\citep{hu2020OGB}. 
The statistics are summarized in \ref{tab: graph_prop_data}.

% \begin{table}[!tp]\centering
% \caption{Statistics of graph classification datasets.}\label{tab: graph_class_data}
% \scriptsize
% \begin{tabular}{lcccccccccc}\toprule
% &DD &Enzymes &Proteins &NCI &IMDB-B &IMDB-M &Reddit-B &Reddit-12K &Collab \\
% \cmidrule(lr){2-4} \cmidrule(lr){5-5} \cmidrule(lr){6-9} \cmidrule(lr){10-10}
% Domain &\multicolumn{3}{c}{Biology} &Chemical &\multicolumn{4}{c}{Social} &Academic \\
% \midrule
% \#Graphs &1,178 &600 &1,113 &4,110 &1,000 &1,500 &2,000 &11,929 &5,000 \\
% \#Classes &2 &6 &2 &2 &2 &3 &2 &11 &3 \\
% \#Nodes &284 &33 &40 &30 &20 &13 &430 &391 &74 \\
% \#Edges &716 &64 &73 &32 &97 &66 &498 &457 &2458 \\
% \bottomrule
% \end{tabular}
% \end{table}

\begin{table}[!htp]\centering
% \vspace{-2em}
\caption{Statistics of graph property prediction datasets.}\label{tab: graph_prop_data}
\scriptsize
% \resizebox{0.5\textwidth}{!}{
\begin{tabular}{lccccccc}\toprule
Domain &Dataset &Task type &\# Graphs &\# Tasks &\# Nodes &\# Edges \\
\midrule
\multirow{3}{*}[-0.8em]{Biology} &DD &Classification &1,178 &2 &284 &716 \\
\cmidrule(lr){2-7}
&Enzymes &Classification &600 &6 &33 &64 \\
\cmidrule(lr){2-7}
&Proteins &Classification &1,113 &2 &40 &73 \\
\midrule
Academic &Collab &Classification &5,000 &3 &74 &2458 \\
\midrule
\multirow{4}{*}[-1em]{Social} &IMDB-B &Classification &1,000 &2 &20 &97 \\
\cmidrule(lr){2-7}
&IMDB-M &Classification &1,500 &3 &13 &66 \\
\cmidrule(lr){2-7}
&Reddit-5k &Classification &4,999 &5 &509 &595 \\
\cmidrule(lr){2-7}
&Reddit-12k &Classification &11,929 &11 &391 &1305 \\
\midrule
\multirow{4}{*}[-1em]{Chemical} &NCI &Classification &4,110 &2 &30 &32 \\
\cmidrule(lr){2-7}
&ogbg-Lipo &Regression &4200 &1 &27 &59 \\
\cmidrule(lr){2-7}
&ogbg-ESOL &Regression &1128 &1 &13 &27 \\
\cmidrule(lr){2-7}
&ogbg-FreeSolv &Regression &642 &1 &9 &17 \\
\bottomrule
\end{tabular}
% }
% \vspace{-2em}
\end{table}

\textbf{Link prediction datasets} include Cora, Citeseer, and Pubmed~\citep{sen2008collective}, Power~\citep{watts1998collective}, Yeast~\citep{bu2003topological}, Erdos~\citep{pajekdatasets}, Amazon Photo~\citep{shchur2018pitfalls}, and Flickr~\citep{snapnets}.
The statistics are summarized in \ref{tab: link_pred_data}.

\begin{table}[!htp]\centering
\caption{Statistics of link prediction datasets.}\label{tab: link_pred_data}
\scriptsize
\begin{tabular}{lccccccccccc}\toprule
&Cora &Citeseer &Pubmed &Power &YST &ERD &Flickr \\
\cmidrule(lr){2-4} \cmidrule(lr){5-5} \cmidrule(lr){6-6} \cmidrule(lr){7-7} \cmidrule(lr){8-8}
Domain &\multicolumn{3}{c}{Citation} &Transport &Biology &Academic &Social \\
\midrule
\#Nodes &2,708 &3,327 &18,717 &4,941 &2,284 &6,927 &334,863 \\
\#Edges &5,278 &4,676 &44,327 &6,594 &6,646 &11,850 &899,756 \\
Mean Degree &3.9 &2.81 &4.74 &2.67 &5.82 &3.42 &5.69 \\
\bottomrule
\end{tabular}
\end{table}

\textbf{Node classification datasets} include Cora, Citeseer, and Pubmed~\citep{sen2008collective}, WebKB (Texas, Cornell, and Wisconsin)~\citep{Pei2020GeomGCN}, Wikipedia Network (Chameleon and Squirrel)~\citep{Pei2020GeomGCN}, and Actor~\citep{tang2009social}.
The first three are homophilic graphs, and the others are heterophilic.
The statistics are summarized in \ref{tab: node_class_data}.

\begin{table}[!htp]\centering
\begin{threeparttable}
\caption{Statistics of node classification datasets.}\label{tab: node_class_data}
\scriptsize
\begin{tabular}{lcccccccccc}\toprule
&Cora &Citeseer &Pubmed &Cornell &Wisconsin &Texas &Chameleon* &Squirrel* &Actor \\
% \cmidrule(lr){2-4} \cmidrule(lr){5-10}
% Homophily &\multicolumn{3}{c}{Homophilic} &\multicolumn{6}{c}{Heterophilic} \\
\cmidrule(lr){2-4} \cmidrule(lr){5-9} \cmidrule(lr){10-10} 
Domain &\multicolumn{3}{c}{Citation} &\multicolumn{5}{c}{Web} &Social \\
\midrule
\#Nodes &2,708 &3,327 &19,717 &183 &251 &183 &890 &2,223 &7,600 \\
\#Edges &5,278 &4,676 &44,324 &295 &499 &309 &8,854 &46,998 &33,544 \\
\#Classes &7 &6 &3 &5 &5 &5 &5 &5 &5 \\
\bottomrule
\end{tabular}
\begin{tablenotes}
  \small
  \item *Chameleon and Squirrel are filtered to remove duplicated nodes~\citep{platonov2023a}.
\end{tablenotes}
\end{threeparttable}
\end{table}

\section{Experiment}
\label{app: experiment}
In this section, we introduce the implementation details and additional results for the experiments.
Throughout all the experiments, we train all the methods with Adam optimizer on an A100 GPU.
We train the guidance head of \method with cross-entropy loss for class labels and mean squared error loss for all other objectives.
For multi-class objectives, we apply the label smoothing~\citep{szegedy2016rethinking} technique following NOS~\citep{gruver2024protein}. Denote $y$ as the one-hot label and $C$ as the number of classes, we have
\begin{equation}
    \overline{\mathbf{y}}_{t}=\bar{\alpha}_{t} * \mathbf{y}+\left(1-\bar{\alpha}_{t}\right) / C * \mathbf{1}.
\end{equation}

\subsection{Graph property prediction}
\label{app: exp_graph_prop_pred}
For graph classification, we follow \citep{Errica2020A} for the setting with $10$-fold cross-validation.
We utilize a $5$-layer GIN with latent dimensions of $64$ throughout the datasets.
For molecule regression, we implement a $5$-layer GIN with a virtual node, and the latent dimensions are $300$.
We have mainly four hyperparameters for \method: step-size $\gamma$ and regularization strength $\lambda$ in \eqref{eq: guidance}, number of repeats per training graph, and whether augment validation and test graphs with the trained guidance head.
For each training graph, we repeatedly generate structures and plug in the original node features for multi-repeat augmentation.
We perform the update in \eqref{eq: guidance} for $5$ times per each sampling step.
The hyperparameters are tuned from the choices in Table~\ref{tab: graph_params}.

\begin{table}[!htp]\centering
\caption{Hyperparameter choices for graph property prediction.}\label{tab: graph_params}
\scriptsize
\begin{tabular}{lcc}\toprule
% Hyperparameter & Choices \\
% \cmidrule(lr){2-2}
$\lambda$ &0.01 \\
$\gamma$ &[0.1, 0.5, 1.0] \\
\# repeats &[1, 5, 10, 32, 64] \\
Aug val and test &[True, False] \\
\bottomrule
\end{tabular}
\end{table}

In Section~\ref{sec: main_exp}, we aim to benchmark the capability of cross-domain pre-training of different methods on the same set of pre-training graphs. 
While the pre-training graphs contain vastly different features, we have to align the feature space to allow pre-training for the baseline methods.
There are two ways to tackle the feature heterogeneity issues in the existing literature.
One line of them utilizes LLMs to align text-space graphs~\citep{chen2024text}, which is not applicable to broader classes of graphs.
Other works, like GCOPE~\citep{zhao2024all}, perform dimension reduction to align the feature dimension of different graphs.
We emphasize that dimension reduction methods fail to deal with extreme cases like missing features. 
This phenomenon is pretty common in real life, as a large proportion of the graphs in the Network Repository do not have corresponding features.
Therefore, we simply use the node degrees as the features in Section~\ref{sec: main_exp}.

We understand that removing the node features may result in a performance drop for the baseline methods. 
Note that most of the baselines follow the pre-training paradigm of~\citep{hu2019strategies} with domain-specific model designs for chemistry and biology datasets, and thus cannot be directly applied to the chosen graph classification datasets. 
Therefore, we adapt the \textbf{semi-supervised}~\citep{you2020l2} and \textbf{self-supervised}~\citep{Sun2020InfoGraph} setting for the baselines for a comprehensive benchmark. 
The semi-supervised setting involves pre-training with all data of that specific dataset and finetuning the training set of each split. 
Meanwhile, baselines of the self-supervised setting pre-train on the whole dataset and then classify the learned graph embeddings with a downstream SVM classifier. 
The results are summarized in Table~\ref{tab: graph_class_comparison}, where the best and second-best results are highlighted in \textbf{bold} and \textit{italic}, respectively.
We observe that UniAug presents consistently satisfactory performance according to the average rank, matching or outperforming the best baseline.

\begin{table}[!htp]\centering
\caption{
\small
Mean and standard deviation of accuracy (\%) with $10$-fold cross-validation on graph classification. The best and second-best results are highlighted in \textbf{bold} and \textit{italic}. The last column is the average rank.}\label{tab: graph_class_comparison}
\scriptsize
\resizebox{\textwidth}{!}{
\begin{tabular}{lccccccccc}\toprule
& &DD &Proteins &NCI1 &IMDB-B &IMDB-M &Reddit-B &Collab &A.R. \\
\cmidrule{2-10}
\multirow{4}{*}{Semi-supervised} &CtxtPred &74.66±0.51 &70.23±0.63 &73.00±0.30 &-- &-- &88.66±0.95 &73.69±0.37 &6.80 \\
&InfoMax &75.78±0.34 &72.27±0.40 &74.86±0.26 &-- &-- &88.66±0.95 &73.76±0.29 &5.60 \\
&GraphCL &\textit{76.17±1.37} &74.17±0.34 &74.63±0.25 &-- &-- &89.11±0.19 &74.23±0.21 &4.60 \\
&JOAO &75.81±0.73 &73.31±0.48 &74.86±0.39 &-- &-- &88.79±0.65 &75.53±0.18 &4.60 \\
\midrule
\multirow{4}{*}{Self-supervised} &InfoGraph &-- &74.44±0.31 &76.20±1.06 &73.03±0.87 &49.69±0.53 &82.50±1.42 &70.65±1.13 &5.17 \\
&GraphCL &-- &74.39±0.45 &77.87±0.41 &71.14±0.44 &48.58±0.67 &\textit{89.53±0.84} &71.36±1.15 &4.50 \\
&JOAO &-- &74.55±0.41 &78.07±0.47 &70.21±3.08 &49.20±0.77 &85.29±1.35 &69.50±0.36 &5.17 \\
&GraphMAE &-- &\textit{75.30±0.39} &\textit{80.40±0.30} &\textbf{75.52±0.66} &\textbf{51.63±0.52} &88.01±0.19 &\textbf{80.32±0.46} &2.17 \\
\midrule
&UniAug &\textbf{78.13±2.61} &\textbf{75.47±2.50} &\textbf{80.54±1.77} &\textit{73.50±2.48} &\textit{50.13±2.05} &\textbf{92.28±1.59} &\textit{77.00±2.02} &1.43 \\
\bottomrule
\end{tabular}
}
\end{table}

To demonstrate the effectiveness of \method in scenarios with limited labeled data, we perform 5-shot graph classification following \citep{liu2023graphprompt}. 
The results are summarized in Table~\ref{tab: graph_class_5shot}
These results show that \method achieves significant performance improvements over the self-supervised baselines, underscoring its robustness and adaptability in few-shot settings.

\begin{table}[!htp]\centering
\caption{Accuracy of 5-shot graph classification.}\label{tab: graph_class_5shot}
\scriptsize
\begin{tabular}{lccc}\toprule
&Proteins &Enzymes \\
\cmidrule{2-3}
GIN &58.17 ± 8.58 &20.34 ± 5.01 \\
\midrule
InfoGraph &54.12 ± 8.20 &20.90 ± 3.32 \\
GraphCL &56.38 ± 7.24 &28.11 ± 4.00 \\
JOAO &57.21 ± 6.91 &35.31 ± 3.79 \\
GraphMAE &60.03 ± 5.35 &33.91 ± 6.58 \\
\midrule
UniAug &\textbf{66.85 ± 4.71} &\textbf{48.37 ± 4.77} \\
\bottomrule
\end{tabular}
\end{table}

In addition, to showcase the flexibility of \method on the downstream backbone, we pick one of the SOTA method PIN~\citep{truong2024weisfeiler} for graph classification and evaluate \method on the basis of it.
The results are summarized in Table~\ref{tab: graph_class_pin}, where we see \method offers constant improvements over PIN.

\begin{table}[!htp]\centering
\caption{Accuracy of graph classification with PIN.}\label{tab: graph_class_pin}
\scriptsize
\begin{tabular}{lccccc}\toprule
&Proteins &NCI1 &IMDB-B \\
\cmidrule{2-4}
PIN &78.8 ± 4.4 &85.1 ± 1.5 &76.6 ± 2.9 \\
UniAug - PIN &\textbf{80.2 ± 2.8} &\textbf{86.5 ± 1.4} &\textbf{77.9 ± 1.8} \\
\bottomrule
\end{tabular}
\end{table}

\subsection{Link prediction}
\label{app: link_pred}
For link prediction, we follow the model designs and evaluation protocols of \citep{li2024evaluating}.
For results based on GCN and NCN, we use a GCN encoder to produce node embeddings and perform link prediction with a prediction head.
The prediction head of GCN is a $3$-layer MLP. 
The number of layers and the latent dimension of the GCN encoder are taken from \citep{li2024evaluating}.
We have mainly three hyperparameters for \method: step-size $\gamma$ and regularization strength $\lambda$, and the number of updates in \eqref{eq: guidance} per each sampling step.
In addition, inspired by the pseudo labeling strategy~\citep{botao2023deep}, we provide an option threshold $q$ for the sampling process of the diffusion model.
Specifically, we only keep the edges with the probability of existence higher than $q$ for each sampling step.
After the sampling process, we recover the training edges of the original graph structure.
The hyperparameters are tuned from the choices in Table~\ref{tab: link_params}.
One thing to mention is that we handle the large graphs by graph partitioning with METIS~\citep{METIS}.
Specifically, we augment the partitions of a large graph and then assemble the partitions back into a single graph.
The edges between different partitions are recovered after the assembling process.

\begin{table}[!htp]\centering
\caption{Hyperparameter choices for link prediction.}\label{tab: link_params}
\scriptsize
\begin{tabular}{lcc}\toprule
$\lambda$ &[0.01, 1, 100] \\
$\gamma$ &[0.1, 1.0, 10.0] \\
$q$ &[None, 0.9, 0.99, 0.999] \\
\# updates &[5, 10, 20] \\
\bottomrule
\end{tabular}
\end{table}

\begin{table*}[!htp]\centering
% \vspace{-2em}
\caption{
\small
Mean and standard deviation across $10$ runs on link prediction. Results are scaled $\times 100$. The last two methods are based on NCN, while the rest are GCN-based. The best result is \textbf{bold} for two backbones, respectively. The \colorbox{lightgray}{highlighted} results indicate negative transfer for pre-training methods compared to GCN. The last column is the average rank of each GCN-based method.}\label{tab: link_pred}
% \vspace{-1em}
\scriptsize
\resizebox{0.8\textwidth}{!}{
\begin{tabular}{lcccccccc}\toprule
&Cora &Citeseer &Pubmed &Power &Yeast &Erdos &Flickr &\multirow{2}{*}{A.R.} \\
&MRR &MRR &MRR &Hits@10 &Hits@10 &Hits@10 &Hits@10 & \\
\cmidrule{2-8}
GCN &30.26 ± 4.80 &50.57 ± 7.91 &16.38 ± 1.30 &30.61 ± 4.07 &24.71 ± 4.92 &35.71 ± 2.65 &8.10 ± 2.58 &4.14 \\
\midrule
AttrMask &\cellcolor[HTML]{d9d9d9}13.43 ± 1.93 &\cellcolor[HTML]{d9d9d9}20.23 ± 1.29 &16.39 ± 3.62 &\cellcolor[HTML]{d9d9d9}29.92 ± 2.61 &25.10 ± 4.77 &\cellcolor[HTML]{d9d9d9}30.85 ± 3.13 &8.77 ± 1.65 &6.43 \\
CtxtPred &\cellcolor[HTML]{d9d9d9}15.68 ± 2.91 &\cellcolor[HTML]{d9d9d9}22.31 ± 1.31 &\cellcolor[HTML]{d9d9d9}13.10 ± 3.70 &\cellcolor[HTML]{d9d9d9}29.30 ± 3.55 &\cellcolor[HTML]{d9d9d9}22.96 ± 4.28 &\cellcolor[HTML]{d9d9d9}34.82 ± 2.55 &\cellcolor[HTML]{d9d9d9}3.61 ± 1.01 &7.86 \\
EdgePred &\cellcolor[HTML]{d9d9d9}15.31 ± 3.54 &\cellcolor[HTML]{d9d9d9}22.91 ± 1.87 &17.85 ± 4.45 &\cellcolor[HTML]{d9d9d9}29.54 ± 3.78 &25.78 ± 4.51 &\cellcolor[HTML]{d9d9d9}34.65 ± 3.84 &\cellcolor[HTML]{d9d9d9}6.86 ± 3.24 &5.43 \\
InfoMax &\cellcolor[HTML]{d9d9d9}16.35 ± 2.57 &\cellcolor[HTML]{d9d9d9}22.90 ± 1.30 &\cellcolor[HTML]{d9d9d9}15.91 ± 2.71 &\cellcolor[HTML]{d9d9d9}29.29 ± 4.72 &26.33 ± 4.12 &35.82 ± 4.12 &\cellcolor[HTML]{d9d9d9}3.23 ± 0.38 &6.00 \\
JOAO &\cellcolor[HTML]{d9d9d9}17.21 ± 3.66 &\cellcolor[HTML]{d9d9d9}23.10 ± 1.41 &\cellcolor[HTML]{d9d9d9}15.33 ± 3.70 &\cellcolor[HTML]{d9d9d9}28.98 ± 4.01 &26.47 ± 4.65 &\cellcolor[HTML]{d9d9d9}33.77 ± 3.05 &\cellcolor[HTML]{d9d9d9}6.01 ± 1.57 &6.00 \\
D-SLA &\cellcolor[HTML]{d9d9d9}15.55 ± 3.12 &\cellcolor[HTML]{d9d9d9}23.05 ± 1.54 &16.10 ± 3.96 &\cellcolor[HTML]{d9d9d9}29.37 ± 2.88 &26.15 ± 3.32 &36.02 ± 4.58 &\cellcolor[HTML]{d9d9d9}6.70 ± 2.03 &5.29 \\
GraphMAE &\cellcolor[HTML]{d9d9d9}15.94 ± 1.73 &\cellcolor[HTML]{d9d9d9}20.35 ± 1.52 &\cellcolor[HTML]{d9d9d9}13.80 ± 1.36 &\cellcolor[HTML]{d9d9d9}27.69 ± 1.99 &26.51 ± 2.92 &35.63 ± 3.61 &8.41 ± 2.44 &6.14 \\
\midrule
CFLP &33.62 ± 6.44 &\textbf{55.20 ± 4.16} &17.01 ± 2.75 &16.02 ± 8.31 &24.23 ± 5.23 &28.74 ± 2.38 & OOM &6.43 \\
\midrule
\method -GCN &\textbf{35.36 ± 7.88} &54.66 ± 4.55 &\textbf{17.28 ± 1.89} &\textbf{34.36 ± 1.68} &\textbf{27.52 ± 4.80} &\textbf{39.67 ± 4.51} &\textbf{9.46 ± 1.18} &1.29 \\
\midrule
NCN &31.72 ± 4.48 &58.03 ± 3.45 &38.26 ± 2.56 &27.36 ± 5.00 &39.85 ± 5.07 &36.81 ± 3.29 &8.33 ± 0.92 & -- \\
\method -NCN &\textbf{35.92 ± 7.85} &\textbf{61.69 ± 3.21} &\textbf{40.30 ± 2.53} &\textbf{30.20 ± 1.46} &\textbf{42.11 ± 5.74} &\textbf{39.26 ± 2.84} &\textbf{8.85 ± 0.90} & -- \\
% NCN
\bottomrule
\end{tabular}
}
% \vspace{-2em}
\end{table*}

In addition to the results shown in Table~\ref{tab: link_pred_comp}, we have the pre-training baselines as mentioned in Section~\ref{sec: main_exp}. 
The results are summarized in Table~\ref{tab: link_pred}.
We observe that existing pre-training methods provide negative transfer, especially on datasets with node features.
More explanation on removing the node features can be found in Appendix~\ref{app: exp_graph_prop_pred}.

\begin{table}[!htp]\centering
\caption{Effects of different guidance objectives.}\label{tab: link_pred_guide}
\scriptsize
\begin{tabular}{lccccccc}\toprule
&Cora &Citeseer &Pubmed &Power &Yeast &Erdos &Flickr \\
&MRR &MRR &MRR &Hits@10 &Hits@10 &Hits@10 &Hits@10 \\
\midrule
Link guide &30.45 ± 2.90 &\textbf{54.66 ± 4.55} &16.97 ± 0.92 &33.41 ± 2.95 &25.80 ± 4.10 &36.79 ± 1.98 &\textbf{9.46 ± 1.18} \\
Degree guide &32.73 ± 6.71 &51.13 ± 5.51 &16.37 ± 0.58 &32.88 ± 2.02 &\textbf{27.52 ± 4.80} &\textbf{39.67 ± 4.51} &9.11 ± 0.88 \\
CN guide &\textbf{35.36 ± 7.88} &50.86 ± 5.73 &\textbf{17.28 ± 1.89} &\textbf{34.36 ± 1.68} &26.67 ± 4.02 &36.18 ± 4.32 &9.28 ± 1.18 \\
\bottomrule
\end{tabular}
\end{table}

As mentioned in Section~\ref{sec: main_exp}, we choose three guidance objectives for link prediction with different granularity.
The effects of different objectives can be found in Table~\ref{tab: link_pred_guide}.
We observe that the outcomes of different objectives differ across datasets and there is no consistently winning strategy.

\subsection{Node classification}
\label{app: node_class}
For node classification on heterophilic graphs, we use the fixed splits from Geom-GCN~\citep{Pei2020GeomGCN} for Cornell, Wisconsin, Texas, and Actor.
For Chameleon and Squirrel, we remove duplicated nodes following \citep{platonov2023a} and take their fixed splits.
Regarding node classification on homophilic graphs, we employ the semi-supervised setting~\citep{yang2016revisiting}.
The GCN backbone is implemented as a $2$-layer classifier.
Similar to graph property prediction, we have mainly four hyperparameters for \method: step-size $\gamma$ and regularization strength $\lambda$ in \eqref{eq: guidance}, number of repeats per training graph, and whether augment validation and test graphs with the trained guidance head.
The hyperparameters are tuned from the choices in Table~\ref{tab: node_params}.

\begin{table}[!htp]\centering
\caption{Hyperparameter choices for node classification.}\label{tab: node_params}
\scriptsize
\begin{tabular}{lcc}\toprule
% Hyperparameter & Choices \\
% \cmidrule(lr){2-2}
$\lambda$ &0.01 \\
$\gamma$ &[0.1, 0.5, 1.0] \\
\# repeats &[1, 5, 10] \\
Aug val and test &[True, False] \\
\bottomrule
\end{tabular}
\end{table}

\begin{table*}[!htp]\centering
% \vspace{-0em}
\caption{
\small
Mean and standard deviation of accuracy (\%) across $10$ splits on node classification of heterophilic graphs. The best result is \textbf{bold}. The \colorbox{lightgray}{highlighted} results indicate negative transfer for pre-training methods compared to GCN. The last column is the average rank of each method.}\label{tab: node_class_hete}
% \vspace{-1em}
% \resizebox{0.5\textwidth}{!}{
\begin{threeparttable}
% \scriptsize
\resizebox{0.8\textwidth}{!}{
\begin{tabular}{lcccccccccc}\toprule
&Cornell &Wisconsin &Texas &Actor &Chameleon* &Squirrel* &A.R. \\
\cmidrule{2-7}
GCN &59.41 ± 6.03 &51.68 ± 4.34 &63.78 ± 4.80 &30.58 ± 1.29 &40.94 ± 3.91 &39.11 ± 1.74 &3.83 \\
\midrule
AttrMask &\cellcolor[HTML]{d9d9d9}44.86 ± 5.43 &53.73 ± 4.31 &\cellcolor[HTML]{d9d9d9}60.54 ± 5.82 &\cellcolor[HTML]{d9d9d9}25.31 ± 1.03 &\cellcolor[HTML]{d9d9d9}35.81 ± 2.88 &\cellcolor[HTML]{d9d9d9}30.63 ± 1.68 &5.83 \\
CtxtPred &\cellcolor[HTML]{d9d9d9}40.81 ± 7.78 &\cellcolor[HTML]{d9d9d9}36.67 ± 17.23 &\cellcolor[HTML]{d9d9d9}58.92 ± 4.32 &\cellcolor[HTML]{d9d9d9}23.97 ± 2.63 &\cellcolor[HTML]{d9d9d9}24.36 ± 4.13 &\cellcolor[HTML]{d9d9d9}26.26 ± 7.50 &9.50 \\
EdgePred &\cellcolor[HTML]{d9d9d9}42.70 ± 5.51 &\cellcolor[HTML]{d9d9d9}48.04 ± 6.63 &\cellcolor[HTML]{d9d9d9}59.37 ± 5.11 &\cellcolor[HTML]{d9d9d9}22.99 ± 6.22 &\cellcolor[HTML]{d9d9d9}21.02 ± 5.06 &\cellcolor[HTML]{d9d9d9}27.94 ± 8.41 &8.83 \\
InfoMax &\cellcolor[HTML]{d9d9d9}39.19 ± 12.75 &\cellcolor[HTML]{d9d9d9}39.80 ± 16.38 &\cellcolor[HTML]{d9d9d9}58.87 ± 4.06 &\cellcolor[HTML]{d9d9d9}23.30 ± 4.37 &\cellcolor[HTML]{d9d9d9}22.59 ± 4.91 &\cellcolor[HTML]{d9d9d9}27.52 ± 9.09 &10.17 \\
JOAO &\cellcolor[HTML]{d9d9d9}40.13 ± 8.60 &\cellcolor[HTML]{d9d9d9}44.70 ± 7.45 &\cellcolor[HTML]{d9d9d9}57.06 ± 3.43 &\cellcolor[HTML]{d9d9d9}24.17 ± 5.02 &\cellcolor[HTML]{d9d9d9}25.81 ± 3.79 &\cellcolor[HTML]{d9d9d9}31.72 ± 7.03 &8.33 \\
D-SLA &\cellcolor[HTML]{d9d9d9}41.05 ± 6.88 &\cellcolor[HTML]{d9d9d9}42.13 ± 9.58 &\cellcolor[HTML]{d9d9d9}59.93 ± 4.29 &\cellcolor[HTML]{d9d9d9}23.74 ± 4.06 &\cellcolor[HTML]{d9d9d9}26.49 ± 4.27 &\cellcolor[HTML]{d9d9d9}28.50 ± 6.90 &8.00 \\
GraphMAE &\cellcolor[HTML]{d9d9d9}47.05 ± 4.37 &57.06 ± 4.59 &63.70 ± 5.51 &\cellcolor[HTML]{d9d9d9}24.69 ± 0.68 &\cellcolor[HTML]{d9d9d9}37.18 ± 3.08 &\cellcolor[HTML]{d9d9d9}31.94 ± 1.65 &5.00 \\
\midrule
Half-Hop &62.46 ± 7.58 &76.47 ± 2.61 &72.35 ± 4.27 &33.95 ± 0.68 &38.59 ± 2.89 &37.34 ± 2.18 &3.00 \\
\method &68.11 ± 6.72 &69.02 ± 4.96 &73.51 ± 5.06 &33.11 ± 1.57 &\textbf{43.84 ± 3.39} &\textbf{41.90 ± 1.90} &2.00 \\
\method + Half-Hop &\textbf{72.43 ± 5.81} &\textbf{79.61 ± 5.56} &\textbf{77.03 ± 4.27} &\textbf{34.97 ± 0.55} &41.94 ± 2.77 &38.79 ± 2.61 &1.50 \\
\bottomrule
\end{tabular}
}
\begin{tablenotes}
  \small
  \item *Chameleon and Squirrel are filtered to remove duplicated nodes~\citep{platonov2023a}.
\end{tablenotes}
\end{threeparttable}
% }
% \vspace{-0em}
\end{table*}

In addition to the results shown in Table~\ref{tab: node_class_hete_comp}, we have the pre-training baselines as mentioned in Section~\ref{sec: main_exp}. 
The results are summarized in Table~\ref{tab: link_pred}, with similar scenarios as graph- and link-level tasks.
The existing pre-training methods provide negative transfer when pre-trained on the same data collection as \method.
More explanation on removing the node features can be found in Appendix~\ref{app: exp_graph_prop_pred}.

To showcase the flexibility of \method on the downstream backbone, we pick one of the SOTA methods PolyGCL~\citep{chen2024polygcl} for node classification on heterophilic datasets and evaluate \method on the basis of it.
The results are summarized in Table~\ref{tab: node_class_hete_polygcl}, where we see \method produces improvements over PolyGCL in three out of four datasets.

\begin{table}[!htp]\centering
\caption{Accuracy of node classification with PolyGCL.}\label{tab: node_class_hete_polygcl}
\scriptsize
\begin{tabular}{lccccccc}\toprule
&Cornell &Wisconsin &Texas &Actor \\
\cmidrule{2-5}
PolyGCL &82.62 ± 3.11 &85.50 ± 1.88 &\textbf{88.03 ± 1.80} &41.15 ± 0.88 \\
UniAug - PolyGCL &\textbf{84.31 ± 2.88} &\textbf{88.35 ± 2.58} &86.70 ± 2.77 &\textbf{43.01 ± 1.27} \\
\bottomrule
\end{tabular}
\end{table}

\subsection{Investigation on scaling}
\label{app: scaling}

In Section~\ref {sec: scaling}, we investigate the scaling behavior of \method regarding data scale and pre-training time.
We omit some of the results for a better visualization.
Here we present the numerical results in Table~\ref{tab: data_scaling} and Table~\ref{tab: training_time}.

\begin{figure*}[!htp]
    \centering
    % \vspace{-0.5em}
    \begin{subfigure}[t]{0.42857\textwidth}  % 0.42857
        \centering
        % \resizebox{\textwidth}{!}{
        \includegraphics[width=\textwidth, left]{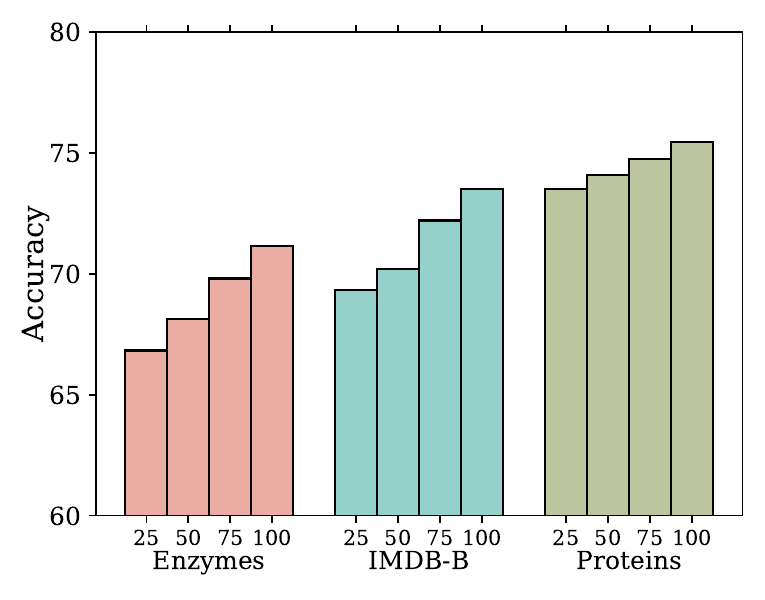}
        % \caption{Lorem ipsum}
    \end{subfigure}%
    ~ 
    \begin{subfigure}[t]{0.57143\textwidth}  % 0.57143
        \centering
        \includegraphics[width=\textwidth, right]{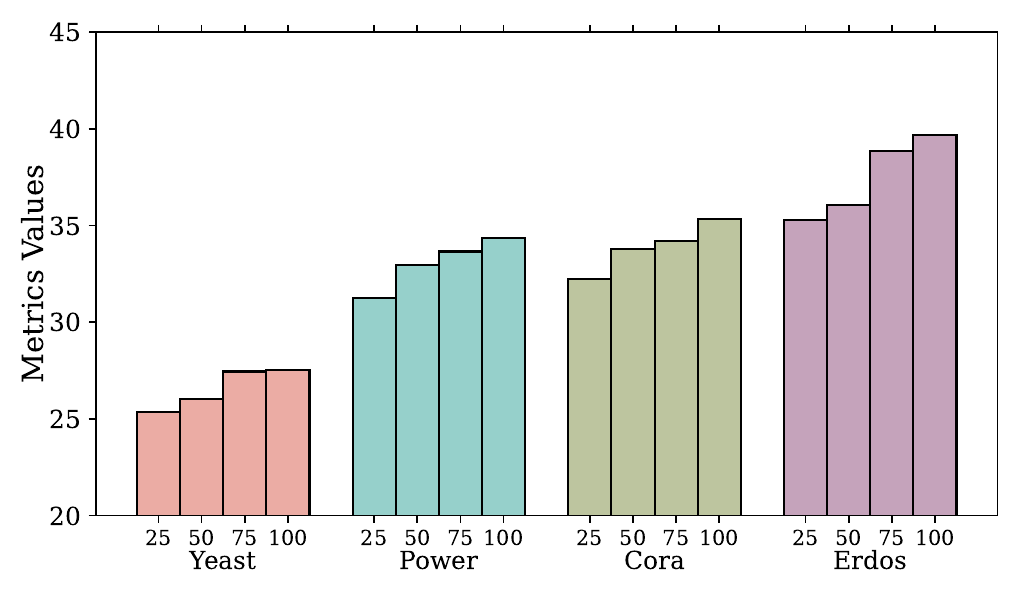}
        % \caption{Lorem ipsum, lorem ipsum,Lorem ipsum, lorem ipsum,Lorem ipsum}
    \end{subfigure}
    % \vspace{-3em}
    \caption{
    \small
    Effects of pre-training data scale (ratio) on graph classification (left) and link prediction (right).
    }
    \label{fig: data_ratio}
    % \vspace{-2em}
\end{figure*}

We recognize that these three sets vary in both scale and diversity. 
To analyze the scaling effect of \method based solely on data quantity, we clustered the pre-training set into 10 clusters based on graph-level representations (Section~\ref{sec: self-cond}) and performed stratified sampling within these clusters. 
From this, we created three subsets containing 25\%, 50\%, and 75\% of the total graphs, and pre-trained \method on each subset.
The results, summarized in Fig.\ref{fig: data_ratio}, show a clear trend of performance improvement as the size of the pre-training set increases. 
Combined with experiments on the SMALL, FULL, and EXTRA sets, these findings suggest that \method benefits from both increasing the scale and enhancing the diversity of the pre-training data.

\begin{table}[!htp]
    \scriptsize
    \caption{
    \small
    Effects of pre-training data scale on graph classification (up) and link prediction (down).
    }
    % \vspace{1em}
    \label{tab: data_scaling}
    \centering
    % \begin{subtable}[h]{0.46\textwidth}
    %     \centering
        % \caption{Accuracy on graph classification.}
    % \resizebox{0.65\textwidth}{!}{
    \begin{tabular}{lccccccc}\toprule
    &Enzymes &Proteins &IMDB-B &IMDB-M \\
    \midrule
    GIN &66.00 ± 7.52 &73.32 ± 4.03 &71.10 ± 2.90 &49.07 ± 2.81 \\
    \midrule
    \method - SMALL &66.83 ± 7.38 &73.50 ± 5.61 &69.80 ± 2.70 &48.93 ± 3.20 \\
    \method - FULL &\textbf{71.33 ± 6.51} &74.05 ± 4.82 &73.11 ± 2.35 &49.67 ± 2.41 \\
    \method - EXTRA &71.17 ± 7.10 &\textbf{75.47 ± 2.50} &\textbf{73.50 ± 2.48} &\textbf{50.13 ± 2.05} \\
    \bottomrule
    \end{tabular}
    % } 
    \\
    \vspace{1em}
    % \end{subtable}
    % \begin{subtable}[h]{0.52\textwidth}
    %     % \caption{Link prediction.}
    %     \scriptsize
    % \resizebox{0.65\textwidth}{!}{
    \begin{tabular}{lcccccccc}\toprule
    &Cora &Citeseer &Power &Yeast &Erdos \\
    &MRR &MRR &Hits@10 &Hits@10 &Hits@10 \\
    \midrule
    GCN &30.26 ± 4.80 &50.57 ± 7.91 &30.61 ± 4.07 &24.71 ± 4.92 &35.71 ± 2.65 \\
    \midrule
    \method - SMALL &32.25 ± 8.71 &47.91 ± 3.87 &32.25 ± 3.72 &25.81 ± 4.89 &36.28 ± 3.56 \\
    \method - FULL &32.81 ± 7.44 &48.32 ± 6.00 &32.97 ± 3.75 &26.36 ± 4.62 &36.07 ± 4.20 \\
    \method - EXTRA &\textbf{35.36 ± 7.88} &\textbf{54.66 ± 4.55} &\textbf{34.36 ± 1.68} &\textbf{27.52 ± 4.80} &\textbf{39.67 ± 4.51} \\
    \bottomrule
    \end{tabular}
    % }
    % \end{subtable}

\end{table}

\vspace{-1em}

\begin{table}[!htp]
    \scriptsize
    \caption{
    \small
    Effects of pre-training amount of compute on graph classification (up) and link prediction (down).
    }
    \vspace{1em}
    \centering
    \label{tab: training_time}
    % \begin{subtable}[h]{0.46\textwidth}
    %     \centering
    %     % \caption{Generated by Spread-LaTeX}\label{tab: }
    %     \scriptsize
    % \resizebox{0.65\textwidth}{!}{
    \begin{tabular}{lcccccc}\toprule
    $10^{-3}$ PF-days &Enzymes &Proteins &IMDB-B &IMDB-M \\
    \midrule
    5 &68.18 ± 6.21 &73.32 ± 3.63 &71.20 ± 2.90 &48.28 ± 2.75 \\
    10 &69.00 ± 5.10 &74.30 ± 5.33 &72.80 ± 3.85 &48.60 ± 2.23 \\
    15 &68.83 ± 5.88 &75.11 ± 3.18 &71.77 ± 2.38 &48.60 ± 2.48 \\
    20 &70.79 ± 5.73 &74.87 ± 5.30 &73.04 ± 2.82 &49.47 ± 2.20 \\
    25 &\textbf{71.50 ± 5.85} &\textbf{75.47 ± 2.50} &\textbf{73.50 ± 2.48} &\textbf{50.13 ± 2.05} \\
    \bottomrule
    \end{tabular}
    % }
    \\
    % \end{subtable}
    \vspace{1em}
    % \begin{subtable}[h]{0.52\textwidth}
    %     \centering
    %     % \caption{Generated by Spread-LaTeX}\label{tab: }
    %     \scriptsize
    % \resizebox{0.65\textwidth}{!}{
    \begin{tabular}{lcccccc}\toprule
    \multirow{2}{*}{$10^{-3}$ PF-days} &Cora &Citeseer &Power &Yeast &Erdos \\
    &MRR &MRR &Hits@10 &Hits@10 &Hits@10 \\
    \midrule
    5 &27.56 ± 4.36 &49.45 ± 9.20 &22.81 ± 9.47 &23.62 ± 9.77 &35.33 ± 3.16 \\
    10 &31.02 ± 6.53 &50.72 ± 6.22 &32.49 ± 2.52 &26.70 ± 4.85 &36.10 ± 4.66 \\
    15 &33.24 ± 7.97 &49.02 ± 5.92 &32.88 ± 3.31 &\textbf{27.80 ± 4.55} &\textbf{39.70 ± 3.67} \\
    20 &34.71 ± 9.08 &52.90 ± 3.84 &33.69 ± 3.23 &26.90 ± 3.93 &39.33 ± 3.16 \\
    25 &\textbf{35.36 ± 7.88} &\textbf{54.66 ± 4.55} &\textbf{34.36 ± 1.68} &27.52 ± 4.80 &39.67 ± 4.51 \\
    \bottomrule
    \end{tabular}
    % }
    % \end{subtable}

\end{table}

% The proposed framework involves a pre-training process

%%%%%%%%%%%%%%%%%%%%%%%%%%%%%%%%%%%%%%%%%%%%%%%%%%%%%%%%%%%%
\newpage
\section*{NeurIPS Paper Checklist}

\begin{enumerate}

\item {\bf Claims}
    \item[] Question: Do the main claims made in the abstract and introduction accurately reflect the paper's contributions and scope?
    \item[] Answer: \answerYes{} % Replace by \answerYes{}, \answerNo{}, or \answerNA{}.
    \item[] Justification: The main claims made in the abstract and introduction accurately reflect the paper's contributions and scope.

\item {\bf Limitations}
    \item[] Question: Does the paper discuss the limitations of the work performed by the authors?
    \item[] Answer: \answerYes{} % Replace by \answerYes{}, \answerNo{}, or \answerNA{}.
    \item[] Justification: It's in the Conclusion and Discussion section.

\item {\bf Theory assumptions and proofs}
    \item[] Question: For each theoretical result, does the paper provide the full set of assumptions and a complete (and correct) proof?
    \item[] Answer: \answerNA{} % Replace by \answerYes{}, \answerNo{}, or \answerNA{}.
    \item[] Justification: The paper does not include theoretical results.

    \item {\bf Experimental result reproducibility}
    \item[] Question: Does the paper fully disclose all the information needed to reproduce the main experimental results of the paper to the extent that it affects the main claims and/or conclusions of the paper (regardless of whether the code and data are provided or not)?
    \item[] Answer: \answerYes{} % Replace by \answerYes{}, \answerNo{}, or \answerNA{}.
    \item[] Justification: The experimental details can be found in the main text and the appendix. We also provide the code in the supplemental material.

\item {\bf Open access to data and code}
    \item[] Question: Does the paper provide open access to the data and code, with sufficient instructions to faithfully reproduce the main experimental results, as described in supplemental material?
    \item[] Answer: \answerYes{} % Replace by \answerYes{}, \answerNo{}, or \answerNA{}.
    \item[] Justification: The data are publicly available and the code is in the supplemental material. 

\item {\bf Experimental setting/details}
    \item[] Question: Does the paper specify all the training and test details (e.g., data splits, hyperparameters, how they were chosen, type of optimizer, etc.) necessary to understand the results?
    \item[] Answer: \answerYes{} % Replace by \answerYes{}, \answerNo{}, or \answerNA{}.
    \item[] Justification: Please see the experiment section and the corresponding appendix.

\item {\bf Experiment statistical significance}
    \item[] Question: Does the paper report error bars suitably and correctly defined or other appropriate information about the statistical significance of the experiments?
    \item[] Answer: \answerYes{} % Replace by \answerYes{}, \answerNo{}, or \answerNA{}.
    \item[] Justification: We report the standard deviation across runs.

\item {\bf Experiments compute resources}
    \item[] Question: For each experiment, does the paper provide sufficient information on the computer resources (type of compute workers, memory, time of execution) needed to reproduce the experiments?
    \item[] Answer: \answerYes{} % Replace by \answerYes{}, \answerNo{}, or \answerNA{}.
    \item[] Justification: See appendix.
    
\item {\bf Code of ethics}
    \item[] Question: Does the research conducted in the paper conform, in every respect, with the NeurIPS Code of Ethics \url{https://neurips.cc/public/EthicsGuidelines}?
    \item[] Answer: \answerYes{} % Replace by \answerYes{}, \answerNo{}, or \answerNA{}.
    \item[] Justification: The research conducted in the paper conforms, in every respect, with the NeurIPS Code of Ethics.

\item {\bf Broader impacts}
    \item[] Question: Does the paper discuss both potential positive societal impacts and negative societal impacts of the work performed?
    \item[] Answer: \answerYes{} % Replace by \answerYes{}, \answerNo{}, or \answerNA{}.
    \item[] Justification: See the impact statement.
    
\item {\bf Safeguards}
    \item[] Question: Does the paper describe safeguards that have been put in place for responsible release of data or models that have a high risk for misuse (e.g., pretrained language models, image generators, or scraped datasets)?
    \item[] Answer: \answerNA{} % Replace by \answerYes{}, \answerNo{}, or \answerNA{}.
    \item[] Justification: The paper poses no such risks.

\item {\bf Licenses for existing assets}
    \item[] Question: Are the creators or original owners of assets (e.g., code, data, models), used in the paper, properly credited and are the license and terms of use explicitly mentioned and properly respected?
    \item[] Answer: \answerYes{} % Replace by \answerYes{}, \answerNo{}, or \answerNA{}.
    \item[] Justification: See appendix.

\item {\bf New assets}
    \item[] Question: Are new assets introduced in the paper well documented and is the documentation provided alongside the assets?
    \item[] Answer: \answerYes{} % Replace by \answerYes{}, \answerNo{}, or \answerNA{}.
    \item[] Justification: See method and experiment sections.

\item {\bf Crowdsourcing and research with human subjects}
    \item[] Question: For crowdsourcing experiments and research with human subjects, does the paper include the full text of instructions given to participants and screenshots, if applicable, as well as details about compensation (if any)? 
    \item[] Answer: \answerNA{} % Replace by \answerYes{}, \answerNo{}, or \answerNA{}.
    \item[] Justification: The paper does not involve crowdsourcing nor research with human subjects.

\item {\bf Institutional review board (IRB) approvals or equivalent for research with human subjects}
    \item[] Question: Does the paper describe potential risks incurred by study participants, whether such risks were disclosed to the subjects, and whether Institutional Review Board (IRB) approvals (or an equivalent approval/review based on the requirements of your country or institution) were obtained?
    \item[] Answer: \answerNA{} % Replace by \answerYes{}, \answerNo{}, or \answerNA{}.
    \item[] Justification: The paper does not involve crowdsourcing nor research with human subjects.

\item {\bf Declaration of LLM usage}
    \item[] Question: Does the paper describe the usage of LLMs if it is an important, original, or non-standard component of the core methods in this research? Note that if the LLM is used only for writing, editing, or formatting purposes and does not impact the core methodology, scientific rigorousness, or originality of the research, declaration is not required.
    %this research? 
    \item[] Answer: \answerNA{} % Replace by \answerYes{}, \answerNo{}, or \answerNA{}.
    \item[] Justification: The core method development in this research does not involve LLMs as any important, original, or non-standard components.

\end{enumerate}

\end{document}